\documentclass{article}
\usepackage{arxiv}

\usepackage[utf8]{inputenc} % allow utf-8 input
\usepackage[T1]{fontenc}    % use 8-bit T1 fonts
\usepackage{hyperref}       % hyperlinks
\usepackage{url}            % simple URL typesetting
\usepackage{booktabs}       % professional-quality tables
\usepackage{amsfonts}       % blackboard math symbols
\usepackage{nicefrac}       % compact symbols for 1/2, etc.
\usepackage{microtype}      % microtypography
\usepackage{lipsum}
\usepackage{fancyhdr}       % header
\usepackage{graphicx}       % graphics
\graphicspath{{media/}}     % organize your images and other figures under media/ folder

\usepackage{bbm}
\usepackage{dsfont}
\usepackage{multirow}
\usepackage{booktabs}

\usepackage{hyperref}
\usepackage{url}
\usepackage{placeins}
\usepackage{booktabs}
\usepackage{siunitx}
\usepackage{threeparttable}
\usepackage{graphicx}
\usepackage{subcaption}
\usepackage{listings}
\usepackage{xcolor}
\usepackage[most]{tcolorbox}
\usepackage{inconsolata}
\usepackage{svg}
\definecolor{promptgreen}{HTML}{4F7A63}
\definecolor{promptbackground}{HTML}{F4F8F5}

\tcbset{
    promptbox/.style={
        enhanced,
        breakable,
        sharp corners,
        boxrule=0.8pt,
        colframe=promptgreen,
        colback=promptbackground,
        colbacktitle=promptgreen,
        coltitle=white,
        fonttitle=\sffamily\bfseries\large,
        left=10pt,
        right=10pt,
        top=12pt,
        bottom=12pt,
        toptitle=6pt,
        bottomtitle=6pt,
        before skip=8pt,
        after skip=8pt,
        listing only,
        listing engine=listings,
        listing options={
            basicstyle=\ttfamily\fontsize{9}{9}\selectfont,
            breaklines=true,
            columns=fixed,
            keepspaces=true,
            showstringspaces=false,
            frame=none
        }
    }
}

\newcommand{\promptfile}[2]{%
    \tcbinputlisting{
        promptbox,
        title={#1},
        listing file={#2}
    }%
}
\title{Reinforcement Learning In Partially Verifiable Environments for Colon Cancer Treatment Planning}

\author{%
  Li-Ching Chen\textsuperscript{1,2},
  Andrés Franco Valiente\textsuperscript{1,3},
  Yuta Ishiyama\textsuperscript{1},
  Zachary R. Chalmers\textsuperscript{1},\\
  \bfseries
  William Chen\textsuperscript{1},
  Kelsey Natsuhara\textsuperscript{1},
  Taema Brinjikji\textsuperscript{1},
  Madhumita Sushil\textsuperscript{1,2},\\
  \bfseries
  Travis Zack\textsuperscript{1,2,4},
  Julian C. Hong\textsuperscript{1,2}\\[2ex]
  {\normalfont
    \textsuperscript{1}University of California, San Francisco\quad
    \textsuperscript{2}Weill Cancer Hub West\quad
    \textsuperscript{3}Vokram Inc.\quad
    \textsuperscript{4}OpenEvidence}
}

\begin{document}

\maketitle

\begin{abstract}
Clinical decision-making is a sequential process of information collection, clinical inference, and forecasting under partial observability, often without a uniquely correct treatment decision. Large language models (LLMs) have demonstrated strong medical question-answering capabilities but extending these advances to clinical reasoning remains challenging because many clinical tasks are only partially verifiable. Recent clinical reinforcement learning (RL) work has therefore largely focused on diagnostic endpoints, where correctness can be more directly verified. Oncology treatment decision making is a particularly challenging instance of this problem that extends beyond medical question-answering in which multiple decision trajectories can be clinically defensible for the same partially observed patient state. We formulate oncology treatment planning as an interactive RL problem under partial observability and construct patient-specific, process-aware rewards from decision processes observed in real world clinical records. The agent must learn to sequentially acquire information from a structured patient representation before committing to a first-line treatment recommendation, and is evaluated on information gathering, treatment selection, and reasoning alignment. Using GRPO and GDPO to optimize Llama-3.1-8B-Instruct within this environment, RL substantially improves the base policy beyond frontier model capabilities. The GDPO-trained policy increases Mean@3 scalar clinical reward by $79.60\%$ over base Llama and exceeds GPT-5.2 by $18.82\%$ under the same interactive environment, while reducing query count during training.  Overall, our results show that patient-specific, partially verifiable rewards can provide an effective learning signal for interactive clinical RL when both the patient state and the underlying clinical decision process are only partially observed.
\end{abstract}

\section{Introduction}
Clinical artificial intelligence has advanced rapidly in its capabilities across clinical reasoning tasks such as medical question-answering \cite{Jin2019-nw, Jin2021-nw, Pal2022-ch}, clinical inference \cite{Tu2025-zt, doi:10.1200/CCI.24.00126, Chen2026-un}, and clinical forecasting \cite{Ma2024-rk}. Existing benchmarks primarily focus on medical question-answering and treatment planning, and often present the model with a curated clinical scenario in which the information relevant to the decision has already been supplied \cite{Kanjee2023-sw}.

Yet many of the most consequential problems in medicine extend beyond identifying a diagnosis to deciding what should be done next when information is only partially observed \cite{Arora2025-jn,Chiu2026-gc,Nori2025-ur,Li2024-hr,Vasilev2025-bb, Schmidgall2026-fe, Jiang2025-px}. Clinical decision-making requires physicians to determine what clinical information is necessary, integrate evidence acquired from multiple sources, and select among interventions whose benefits and risks depend on the particular patient. AI systems intended to support these decisions must therefore do more than map a completed case description to a recommendation; they must participate in the clinical reasoning process through which the information required for a decision is acquired, interpreted and translated into action \cite{Qiu2025-ui, Rao2026-bj}.

% Oncology treatment provides an especially challenging instance of this decision-making problem because treatment decisions depend jointly on the available information about the disease and the longitudinal information about the patient that is acquired throughout the clinical evaluation. AI systems intended to support these decisions must therefore do more than map a completed case description to a recommendation; they must participate in the clinical reasoning process through which the information required for a decision is acquired, interpreted and translated into action.

Even after the relevant evidence has been acquired, there may be no uniquely correct treatment action, with different clinicians reasonably coming to different treatment choices. Multiple prospective treatment plans may remain clinically defensible because treatment decisions reflect tradeoffs among efficacy, toxicity, contraindications, treatment sequencing, patient preferences, feasibility, and other contextual constraints such as institutional regulations and compliance \cite{Krell2015-on}. Thus, evaluating treatment-planning quality requires assessing not only the final recommendation, but also whether the clinical agent acquired and used the information necessary to justify that recommendation \cite{Ruhrberg-Estevez2026-mg}.

The best data available for oncology treatment research is based on real decisions and outcomes within longitudinal clinical records, but even these do not necessarily provide ground-truth annotations of the underlying decision process. In particular, they do not specify which information should be acquired at each intermediate state or which clinical inference should follow from the accumulated evidence. Even at the point of treatment selection, the historically observed conditions need not constitute a unique ground-truth action; different clinician phenotypes can give rise multiple clinically defensible first-line treatment plans for the same patient. In this work, we therefore focus on immediate clinical decision-making: an agent sequentially acquires patient-specific information, performs clinical inference over the acquired evidence, and ultimately selects a first-line treatment. The resulting supervision is therefore naturally expressed as reward over a sequence of decision. 

Our central question is: \textit{how can we design strongly verifiable reinforcement-learning environments when the clinically relevant patient state is partially observed and the available clinical signals provide only partially verifiable rewards?}

Here, we use environment to denote the complete interactive system through which a clinical agent acquires and infers information about the underlying patient state and receives rewards. Most importantly for the setting considered here, environment design determines what aspects of an agent's trajectory can be verified and how those verification signals are translated into rewards. Rather than treating verifiability as binary, we consider rewards along a continuum from weakly to strongly verifiable, according to how unambiguously and reliably the environment can determine whether an agent's behavior satisfies the relevant clinical criteria. 

Our main contributions are threefold.
\begin{enumerate}
    \item We formulate oncology treatment planning as an interactive reinforcement-learning problem under partial observability and construct an environment from retrospective clinical records in which an agent must actively acquire patient-specific information before committing to a treatment recommendation. We couple this interaction with a patient-specific, process-aware reward that evaluates information gathering, treatment selection, and alignment with the documented clinical rationale.

    \item We show that reinforcement learning within this environment can produce a relatively small, specialized language model that substantially outperforms a frontier general-purpose model on the same interactive oncology treatment-planning task. Under the same patient environment, retrieval interface, turn budget, and evaluation procedure, the RL-trained Llama-3.1-8B policy achieves higher treatment-planning reward than GPT-5.2. These results suggest that appropriately constructed domain-specific environments can provide learning signals strong enough for smaller models to acquire specialized sequential decision-making behavior not exhibited by substantially larger models zero-shot.

    \item We provide evidence that the learned policy acquires clinically structured decision-making behavior rather than improving only its final-answer accuracy. During reinforcement learning, the policy improves its overall reward while reducing the number of queries it issues, consistent with learning to acquire more decision-relevant information. Its information-seeking trajectories, treatment rationales, and recommendations are also concordant with the documented decision processes of oncologists. In a preliminary clinician comparison, the specialized policy performs comparably to an oncologist under the same interaction protocol; because this comparison currently contains only a small number of cases, we treat it as suggestive evidence rather than evidence of clinician-level performance. 
\end{enumerate}

\section{Related Work}

\paragraph{Interactive clinical agents and sequential clinical decision-making.} Recent work has extended clinical language models from single-response reasoning to agents that interact sequentially with clinical environments. Similarly, recent clinical RL methods optimize information acquisition. Doctor-R1 and DoctorAgent-RL train agents through multi-turn patient dialogue \cite{Lai2025-xe, Feng2026-pq}. DiagAgent, SALUS and LA-CDM  optimize the policy that seeks information sequentially until it is sufficient confident with the diagnosis \cite{Qiu2025-eu, Gao2026-xu, Bani-Harouni2025-yk}. ResidencyRL broadens this paradigm to long-horizon simulated encounters involving inquiry, tool use, documentation, diagnosis, management, and safety \cite{Lievin2026-ar}.  Most of these environments remain organized around diagnostic endpoints. In contrast, our environment places the patient-specific clinical state itself under partial observability and focuses on improving the agent's reasoning capabilities required to sequentially acquire and interpret evidence from an initially hidden patient record before committing to a first-line treatment decision.

\paragraph{Reinforcement learning under partially verifiable clinical feedback.} Much of the recent reinforcement learning (RL) for reasoning assumes setting in which either final outcome is verifiable or intermediate reasoning steps can be explicitly supervised, such as mathematics and programming \cite{Lightman2023-hu, Jia2025-pi}.  A complementary line of work studies how RL can be extended beyond tasks with deterministic verifiable signals, especially in the clinical setting. Clinical-R1 introduces multi-objective clinical RL that jointly rewards correctness, faithfulness, and comprehensiveness \cite{Gu2025-fk}, while Rubrics as Rewards replaces binary verification with instance-specific, multi-objective feedback based on natural language for on-policy RL \cite{Gunjal2025-tr}. More recently, GatorOnco applies RL to colorectal-cancer treatment planning, using a hierarchical treatment reward together with clinical-guideline retrieval to promote guideline-concordant treatment generation \cite{Lyu2026-dr}.  On the other hand, our reward is anchored to patient-specific decision processes reconstructed from retrospective clinician notes rather than primarily to guideline concordance. Because neither intermediate reasoning actions nor the terminal treatment necessarily admit a unique ground-truth label, we use a terminal but process-aware reward that evaluates the complete interactive trajectory across information acquisition, treatment selection, and clinical reasoning.

% These methods expand the reward space beyond exact-match correctness, but these aforementioned studies hold the clinical prompts fixed instead of allowing for an interactive clinical environment. 

% In our paper, we allow for interactive environments whose reward is given by a multi-objective rubric where the generated clinical prompts are an additional stochastic process. In addition, in order to define proper reward signals for clinical reasoning, we formulate expert reasoning extraction from the retrospective data from the unstructured clinical notes, documenting rationales and information gathered in the clinical encounters, and reconstruct them into coherent reasoning trajectories from the oncologists given patient cases. The process-aware reward model utilizes the expert reasoning data as reward signals when comparing the action trajectories of oncologists with that of the agent.

\section{Clinical Problem Formulation}
First-line treatment planning for patients with colon cancer requires oncologists to integrate disease characteristics, molecular biomarkers, prior treatment history, patient functional status, comorbidities, preference, and treatment intent to determine an appropriate initial management strategy. In clinical practice, decisions are made under incomplete information and unfold through iterative and intertwined phases of data acquisition and clinical inference. Oncologists must identify and acquire the evidence needed to distinguish among potentially defensible treatment options, while continuously interpreting newly acquired evidence to refine subsequent information gathering and candidate treatment plans. These processes are sequential but tightly coupled, iterating until sufficient evidence has been gathered to support a treatment decision. We therefore formulate this process as a stochastic sequential decision-making problem under partial observability (see Figure \ref{fig:workflow}).

% As a result, we provide instructions to the agents to structure their actions in phases, analogous to the clinicians process. These instructions are written in natural language and thus are soft, stochastic constraints which are not deterministically enforced. We provide the agent a one-liner describing the cancer diagnosis and some incomplete information (age, sex) of the patient at the start of the medical question-answering phase in order to mirror the opening of an oncology consultation. Our goal is to then incentivize the agent to acquire the remainder of the patient state by querying the record. We put a hard cap on the number of turns that the agent may take in order to emulate the finite time available in the clinic. 

We reward the agent against the documented per-patient decision trajectory, rather than explicit adherence to published clinical guidelines. We do this in recognition that these guidelines are just that: a starting guide for treatment, but often deviated from actual clinical practice. These deviations due to the nuanced nature of clinical reality are of interest in training models that can replicate the realities of clinical decision making. In addition, the extracted decision trajectory is evidence of one clinically realized decision. There is no unique trace of information-gathering and reasoning process with stepwise label, and for the same patient state, multiple clinically reasonable treatment options can be available. We therefore formulate clinical verifiability as inherently partial: some properties of a trajectory can be evaluated against patient-specific evidence and the documented oncologist decision process, while the overall quality of the trajectory cannot be reduced to agreement with a single uniquely correct sequence of actions. 

\section{Methodology for Clinical Reinforcement Learning Environment Design}

\begin{figure}
    \centering
    \includegraphics[width=\linewidth]{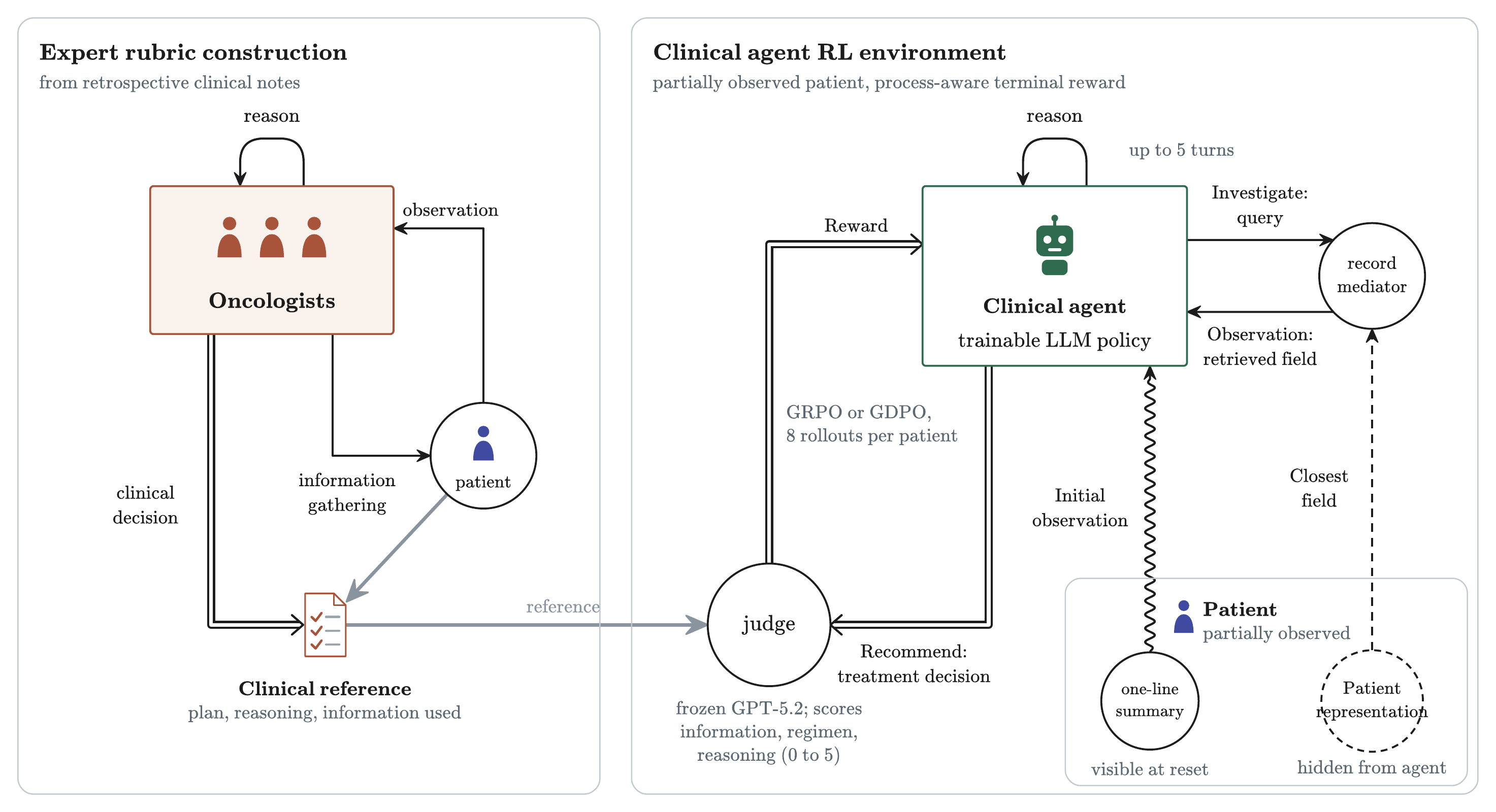}
    \caption{The system consists of five functional components: a patient environment constructed from retrospective clinical data, a trainable clinical agent, a deterministic action parser, a frozen patient-record mediator, and a frozen judge. }
    \label{fig:system}
\end{figure}

% \subsection{System Setup}
Here, we devise an interactive RL environment for colon cancer treatment planning under partial observability and partially verifiable clinical feedback. While what we describe herein is applied to oncology information gathering and decision making, the system is generalizable to RL construct across the broader medical domain.  The system is comprised of a trainable clinical agent interaction with a patient representation through an deterministic action parser and frozen patient-record mediator, while a frozen judge evaluates the completed trajectory. At the start of the trajectory, the agent will receive a one-line summary regarding the patient. The agent provides reasoning and actions between investigation, which is passed through the parser and patient-record mediator to retrieve patient-specific clinical information and return it to the agent, or a terminal treatment recommendation. The completed trajectory is evaluated by the frozen judge using a process-aware reward model, with the resulting terminal reward converted into a group-relative advantage for optimizing the clinical agent's policy. We provide an overview of the system in Figure \ref{fig:system}. We describe each component in detail below, but defer the experimental setup and prompts to Appendices \ref{sec:experiments} and \ref{app:prompt}.

\subsection{Data \& Cohort Construction}
We construct an oncology treatment-planning dataset from retrospective, longitudinal, de-identified electronic health records (EHR)s at the University of California, San Francisco, comprising 950 adult colon cancer patients. We extracted documented first-line treatment plan and their underlying clinical rationales, including the information acquisition process of the oncologists, to supervise our process-aware reward model. To build an interactive environment, we reconstructed longitudinal patient trajectories into a structured representation of history of present illness, oncologic profiles, and physiological data \cite{Chiu2026-gc}. The dataset is split at the patient level into disjoint training, validation, and test sets, comprising 665, 143, and 142 patients. Full details on cohort inclusion and exclusion criteria, patient representation schema definition, and extraction prompts can be found in Appendix \ref{App:Data}.

\subsection{Patient Representation and Environment State}

We index patients by $i$, trajectory by $g$, and interaction turns by  $k \in\{0, \ldots, K-1\}$, where $K$ denotes the turn budget. Let $\chi^i$ denote the underlying clinical state of the patient $i$ and $\rho^i=\rho\left(\ell^i, \mathcal{R}^i\right)$ where $\ell^i$ is a one-line patient summary available at the start of the episode and $\mathcal{R}^i$ is the patient-specific EHR record which is initially hidden but available through the retrieval. Because $\rho^i$ is constructed from documented clinical data, it is an incomplete proxy for the underlying clinical state $\chi^i$.

No clinical time elapses during a trajectory. The agent's actions do not change the patient's physiology or generate counterfactual treatment outcomes. As a result, the environment dynamics arise from the sequential information acquisition, clinical inference and decision planning. 

To facilitate rapid RL training and minimize cost, we converted the unstructured clinical not documenting patient state and clinician decision to a structured key-value dictionary. The retrieval record is represented as  $\mathcal{R}^i=\left\{\left(\xi, v^i(\xi)\right): \xi \in \mathcal{F}^i\right\}$, where  $\mathcal{F}^i$ denote clinical fields available for patient $i$,  $\xi$ is a clinical field and $v^i(\xi)$ is its documented value.  

The formulation of the partially observed Markov decision process will require us to know the fully observed RL trajectory $\omega^{i,g}$ which can be constructed from the fully observed state-action history $H^{i,g}_k = (s_0,a_0,s_0,a_1,...,s_k)$, the patient representation $\rho^i$, the budget $K$ and any additional information flow $x$ that passes through the environment (parser and mediator) which the agent does not observe. The fully observed RL trajectory is then parametrized by
\begin{equation}
\omega_k^{i, g }=\left( H_k^{i, g}, \rho^i, K-k,x\right).
\end{equation}
We leave $x$ as an placeholder symbol for the rest of the environment information that goes to the judge model but the agent does not observe. We further remark that we will formulate the reward model as a terminal reward that does not appear within the history $H^{i,g}_k$ until the terminal step.  Full state and trajectory notation is provided in Appendix~\ref{app:env}.

\subsection{Clinical Agent Action Space, Structured Action Parser \& Frozen Patient-Record Mediator}
At the turn $k$, the clinical agent with a parameterized policy  $\pi_{\theta_n}$ conditioned on the observed transcript generates a natural-language response $a_k^{i,g}$, which denotes the complete token sequence generated at one turn.  

We equip the clinical agent with a deterministic parser that maps each response to either an investigation action or a terminal treatment recommendation. Investigation actions may contain multiple queries, which are screened by deterministic regular expressions to reject requests for clinical inference and retain only requests for clinical variables. Accepted queries are executed independently, while corrective messages are returned for rejected queries. If no valid structured action can be extracted, then no clinical action is executed and the turn is consumed. In order to design the environment in such a way that it is auditable, we equip the agent with a retrieval harness that keeps track of the documents it retrieved. The patient-record mediator maps any accepted action request to the most semantically similar field in the patient representation and returns the associated documented value. We detail the specific of the parser and the mediator in Appendix \ref{app:med}.

We note that the the mediator performs a top-1 retrieval without any abstention threshold. Every accepted request therefore receives a key-value response, even if the intended clinical variable was absent from $\mathcal{F}^i$. In that case, the mediator returns the nearest available field rather than indicating the requested variable was missing. This behavior does introduce semantic retrieval error into the observation channel but it is conditioned on a fixed request, patient representation and the frozen mediator and hence the mapping is deterministic and it is just an approximation error rather than a stochastic observation noise. This error, however can be traced back to its origin because one may always check which clinical field was actually retrieved so auditability is not lost. 

\subsection{Reward Model Construction} \label{sec:reward}
Treatment planning does not admit an exact verifier like in the case of programming or mathematics. Although the treatment selected by an oncologist in our dataset provides a signal of a clinically realized decision, it is not necessarily true that it is the only clinically defensible treatment. 

We therefore evaluate each trajectory against a patient-specific clinician reference $\zeta_{\mathrm{ref}}^i$. The clinician reference contains the documented treatment plan and the corresponding clinical reasoning. The reference is withheld from the clinical agent throughout the interaction with the environment and is supplied only to the judge after termination. We do not provide the agent a reward at each non-terminal turn. Hence, the reward model gives terminal rewards but remains process-aware. 

The judge receives $\left(\omega^{i, g}, \zeta_{\text {ref }}^i, \mathcal{U}\right)$, where $\mathcal{U}$ is a fixed evaluation rubric meant to evaluate the trajectory. The judge then produces three bounded integer scores
\begin{equation}
r_{\text {info },}^{i, g} \quad r_{\text {regimen }}^{i, g} \quad r_{\text {reason }}^{i, g} \in\{0,1, \ldots, 5\}.
\end{equation}
The information-gathering score $r_{\text {info }}^{i, g}$ evaluates whether the clinical agent acquired the information needed for the treatment decision. The score penalizes both premature recommendation and inefficient over-investigation. 

The regimen-accuracy score $r_{\text {regimen }}^{i, g}$ evaluates the terminal treatment recommendation against the clinician reference. Exact agreement receives the highest score, while clinically equivalent alternatives may receive substantial credit. Selecting the wrong treatment class or an incompatible regimen receives substantially less credit than omitting a secondary component. 

The reasoning-alignment score $r^{i,g}_{\text{reason}}$ evaluates whether the agent correctly connects the acquired patient-specific evidence to its recommendation. It measures whether the agent uses relevant variables consistently, interprets them correctly, and appropriately includes or excludes treatment alternatives. This score distinguishes a patient-specific recommendation from an otherwise similar recommendation unsupported by the evidence acquired during the episode.

We define a scalar clinical reward as 
  \begin{equation}
r_{\text {clin }}^{i, g}=\frac{r_{\text {info }}^{i, g}+r_{\text {regimen }}^{i, g}+r_{\text {reason }}^{i, g}}{15} -0.2\cdot\mathbbm{1}_{\text{p}} -0.4\cdot\mathbbm{1}_{\text{c }},
\end{equation} \label{clin}
where $\mathbbm{1}_p$ is an indicator that is nonzero if there is any error message from the parser. $\mathbbm{}{1}_c$ is a completeness indicator that is nonzero if the agent did not output a treatment recommendation before the turns run out. Each indicator is a penalty that can only be activated once per trajectory. 

\subsection{Objectives and Algorithms: GRPO and GDPO}
Group Relative Policy Optimization (GRPO) is suitable for our setting because each patient-specific environment can generate multiple trajectories under the same underlying patient representation \cite{Shao2024-kd}. These trajectories form a rollout group, allowing the relative quality of different information-gathering, reasoning, and treatment trajectories for the same patient to be estimated without learning a separate value function. 

Since GRPO first scalarizes the reward components into one terminal reward, then performs group normalization, and finally calculates the relative advantage for each trajectory within the rollout group, it cannot distinguish two rollouts with different clinical reward profiles but the same scalar total. For example, strong information gathering followed by an incorrect regimen may receive the same scalar reward as weak information gathering followed by an accurate regimen. Group reward-Decoupled Normalization Policy Optimization (GDPO) addresses this loss of reward resolution by normalizing each clinical reward component separately before combining the resulting component advantage \cite{Liu2026-nb}. Our hypothesis is that GDPO will be better able to learn the combined tasks of information gathering and decision making than GRPO due to its ability to distinguish between the reward components during advantage calculation.

For both GRPO and GDPO, we sample $G$ trajectories for each patient and construct the policy update from their group relative advantage derived from the terminal rewards.  The same trajectory-level advantage is broadcast to every agent-generated token in that rollout. Only tokens generated by the clinical agent contribute to the policy-gradient term. We mask the tokens belonging to the system prompt, initial patient observation, parser messages, or mapper response. A full description on the advantage formulation and the objective function is provided in Appendix \ref{app:policy}.

\section{Results}
We evaluate on 142 held-out patients with three generations per case. We report the Mean@3 and Best@3 of the reward calculated using LLM-as-a-judge as described in Section \ref{sec:reward}. Mean@3 denotes the average scalar clinical reward across the three generations per case, while Best@3 averages the highest scalar clinical reward among the three generations per case. 

RL improves all three policy backbones, although the magnitude of improvement varies substantially by backbone. For Llama-3.1-8B-Instruct, GRPO and GDPO increase Mean@3 reward from $0.3388$ to  $0.5712$ and $0.6085$, respectively. Qwen3.5-9B improves from $-0.0142$ to $0.1994$ with GRPO and $0.3429$ with GDPO, while GLM-4-9B-0414 improve from $-0.0163$ to $0.5313$ with GRPO and $0.5105$ with GDPO. Among the evaluated policies, GDPO-trained Llama achieves the highest Mean@3 reward and exceeds GPT-5.2, which attains $0.5121$ (see Table \ref{tab:overall_reward}). 

Reward decomposition shows that RL improves Llama across information gathering, treatment selection, and reasoning relative to its base policy, while its relative advantage over GPT-5.2 is concentrated in treatment selection. GRPO-trained and GDPO-trained Llama achieve treatment rewards of \(3.366\) and \(2.791\), respectively, compared with \(2.702\) for GPT-5.2. In contrast, GPT-5.2 remains stronger on information gathering and reasoning with Mean@3 total scalar rewards of \(2.977\) and \(3.016\), respectively. GRPO-trained Llama achieves \(2.754\) and \(2.688\) on these axes, while GDPO-trained Llama achieves \(2.953\) and \(2.791\) (see Tables \ref{tab:reward_axes_main} and \ref{tab:reward_axes}).

\begin{figure}[h]
    \setlength{\belowcaptionskip}{-2pt}
    \centering

    \begin{subfigure}[h]{0.32\linewidth}
        \centering
        \includegraphics[width=\linewidth]
        {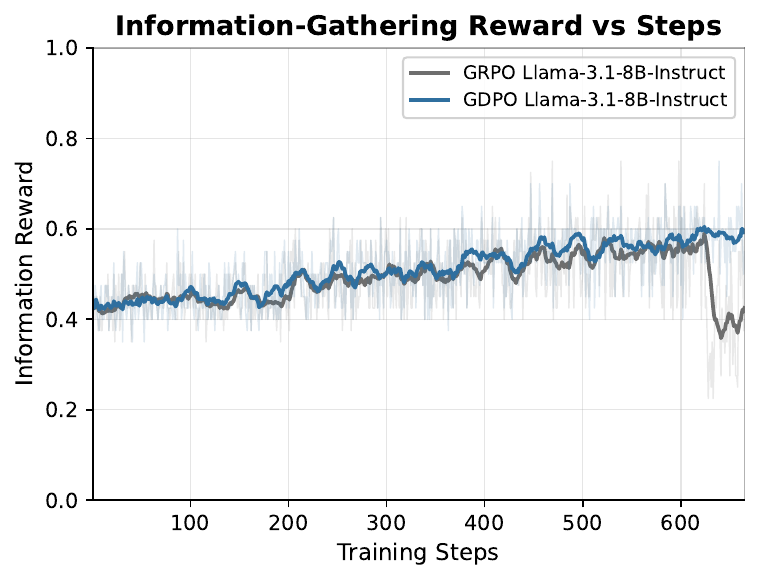}
        \caption{Information-gathering reward.}
        \label{fig:learning_information}
    \end{subfigure}
    \hfill
    \begin{subfigure}[h]{0.32\linewidth}
        \centering
        \includegraphics[width=\linewidth]
        {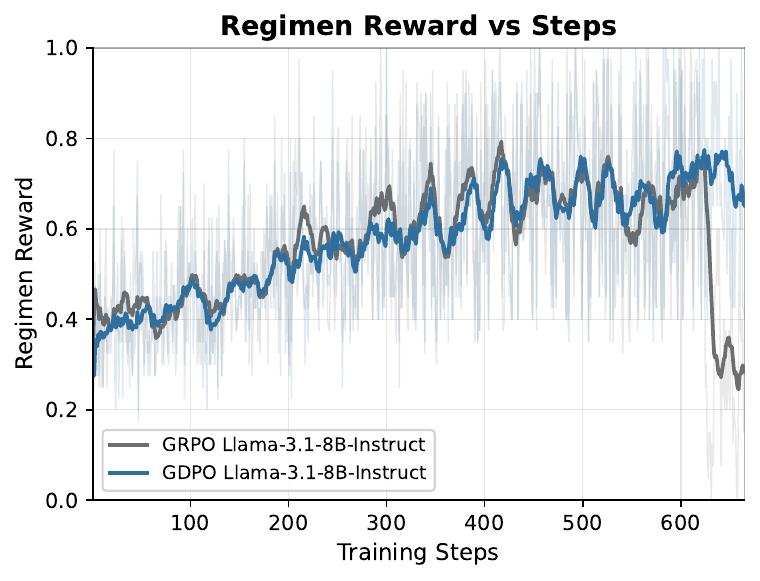}
        \caption{Treatment reward.}
        \label{fig:learning_regimen}
    \end{subfigure}
    \hfill
    \begin{subfigure}[h]{0.32\linewidth}
        \centering
        \includegraphics[width=\linewidth]
        {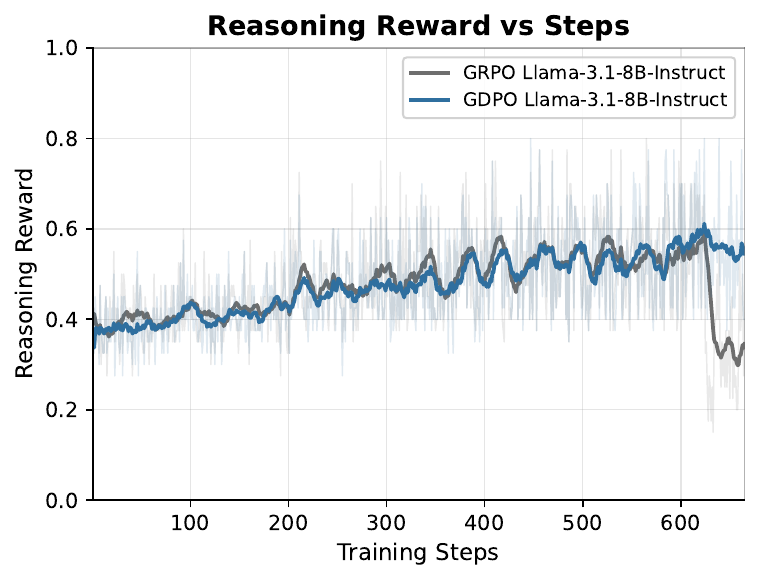}
        \caption{Reasoning reward.}
        \label{fig:learning_reasoning}
    \end{subfigure}

    \caption{
    Learning dynamics of GRPO and GDPO with Llama-3.1-8B-Instruct.
    Dark curves show a 15-step moving average and faint curves show the
    unsmoothed training trajectory. Each curve corresponds to a single training run.
    }
    \label{fig:llama learning_dynamics}
\end{figure}

Across Llama training, treatment reward increases more strongly than information-gathering and reasoning reward, while GRPO exhibits a reward collapse at the later stage of the training as shown in Figure \ref{fig:llama learning_dynamics}. Information-acquisition behavior also changes with RL, although not uniformly across all the policy backbone (see Figures \ref{fig:qwen training} and \ref{fig:glm training}).

To compare the performance of the agent with human, we recruit four oncologists to produce trajectories in the same interactive environment. Each clinician is assigned with 6 to 10 unique cases from the held-out test set. We allow the clinician to send out probe to the patient representation and leave the reasoning in the same fashion as the agent does. Clinicians need to provide a treatment recommendation within 5 turns. We report the average reward across all axes compared to the models on the same patient set. On the same 34 held-out cases performed by four oncologists, GDPO-trained Llama attains an average scalar reward of $0.6000$, compared with $0.573$ for the oncologists, $0.544$ for GPT-5.2, and $0.537$ for GRPO-trained Llama (see Table \ref{tab:clinician_comparison}).

\begin{table}[t]
\centering
\caption{
Overall reward performance across different agent policies.
Results are evaluated on $n=142$ cases with three generations per case.
Mean@3 is the mean scalar reward across three generations; Best@3 is the
mean of the highest scalar reward among the three generations for each case.
Best results are shown in \textbf{bold} and second-best results are \underline{underlined}. 
}
\label{tab:overall_reward}
\small
\setlength{\tabcolsep}{7pt}
\renewcommand{\arraystretch}{1.08}
\begin{tabular}{
    l
    S[table-format=1.4]
    S[table-format=1.4]
    S[table-format=1.4]
    c
}
\toprule

& {Mean@3}
& {Best@3}
& {SD}
& {95\% CI} \\
\midrule
GRPO Llama-3.1-8B-Instruct & \underline{0.5712} & 0.6446 & 0.1465 & [0.547, 0.595] \\
GDPO Llama-3.1-8B-Instruct & \textbf{0.6085} & \textbf{0.6718} & 0.1264 & [0.588, 0.629] \\
GRPO Qwen3.5-9B & 0.1994 & 0.4878 & 0.2096 & [0.165, 0.234] \\
GDPO Qwen3.5-9B & 0.3429 & 0.5808 & 0.2174 & [0.307, 0.379] \\
GRPO GLM-4-9B-0414 & 0.5313 & 0.6263 & 0.1321 & [0.510, 0.553] \\
GDPO GLM-4-9B-0414 & 0.5105 & 0.6192 & 0.1474 & [0.486, 0.535] \\
GPT-5.2 & 0.5121 & \underline{0.6469} & 0.2408 & [0.472, 0.552] \\
Llama-3.1-8B-Instruct & 0.3388 & 0.4934 & 0.1422 & [0.315, 0.362] \\
Qwen3.5-9B & -0.0142 & 0.1474 & 0.1505 & [-0.039, 0.011] \\
GLM-4-9B-0414 & -0.0163 & 0.1981 & 0.1879 & [-0.047, 0.015] \\

\bottomrule
\end{tabular}
\end{table}

\begin{table}[t]
\centering
\caption{
Mean@3 performance on each reward axis.
Information, treatment, and reasoning are scored from 0--5;
completeness and format are scored from 0--1.
Best results are shown in \textbf{bold} and second-best results are \underline{underlined}.
}
\label{tab:reward_axes_main}
\small
\setlength{\tabcolsep}{4.5pt}
\renewcommand{\arraystretch}{1.08}

\begin{tabular}{
    l
    S[table-format=1.3]
    S[table-format=1.3]
    S[table-format=1.3]
    S[table-format=1.3]
    S[table-format=1.3]
}
\toprule
 
{Policy}
& {\shortstack{Information\\{\scriptsize $(0$--$5)$}}}
& {\shortstack{Regimen\\{\scriptsize $(0$--$5)$}}}
& {\shortstack{Reasoning\\{\scriptsize $(0$--$5)$}}}
& {\shortstack{Complete\\{\scriptsize $(0$--$1)$}}}
& {\shortstack{Format\\{\scriptsize $(0$--$1)$}}} \\
\midrule

GRPO Llama-3.1-8B-Instruct
& 2.754 & \underline{3.366} & 2.688 & 0.960 & 1.000 \\

GDPO Llama-3.1-8B-Instruct
& \underline{2.953} & \textbf{3.453} & \underline{2.791} & \textbf{0.988} & 1.000 \\

GRPO Qwen3.5-9B
& 2.606 & 1.453 & 2.298 & 0.439 & 1.000 \\

GDPO Qwen3.5-9B
& 2.725 & 2.033 & 2.399 & 0.664 & 1.000 \\

GRPO GLM-4-9B-0414
& 2.728 & 2.967 & 2.444 & \underline{0.972} & 1.000 \\

GDPO GLM-4-9B-0414
& 2.629 & 2.854 & 2.455 & 0.953 & 1.000 \\

GPT-5.2
& \textbf{2.977} & 2.702 & \textbf{3.016} & 0.831 & 1.000 \\

Llama-3.1-8B-Instruct
& 2.124 & 1.876 & 1.927 & 0.859 & 1.000 \\

Qwen3.5-9B
& 2.406 & 0.437 & 2.085 & 0.143 & 1.000 \\

GLM-4-9B-0414
& 2.002 & 0.678 & 1.723 & 0.225 & 1.000 \\

\bottomrule
\end{tabular}
\end{table}

\setlength{\textfloatsep}{6pt plus 1pt minus 1pt}

\begin{table}[h]
\centering
\caption{
Comparison between oncologists and language-model policies in the same
interactive treatment-planning environment.
Four oncologists each evaluated 6 to 10 unique cases from the held-out test set,
for a total of 34 clinician trajectories.
Agent results are evaluated on the same 34 cases under the same five-turn
interaction protocol and averaged over three generations per case.
Information, treatment, and reasoning rewards are scored from 0--5;
completeness is scored from 0--1.
}
\label{tab:clinician_comparison}
\small
\setlength{\tabcolsep}{5pt}
\renewcommand{\arraystretch}{1.08}

\begin{tabular}{
    l
    S[table-format=2.0]
    S[table-format=1.3]
    S[table-format=1.3]
    S[table-format=1.3]
    S[table-format=1.3]
    S[table-format=1.3]
    S[table-format=2.2]
}
\toprule
\textbf{Policy}
& {$n$}
& {\shortstack{\textbf{Scalar}\\\textbf{Reward}}}
& {\shortstack{\textbf{Information}\\{\scriptsize $(0$--$5)$}}}
& {\shortstack{\textbf{Regimen}\\{\scriptsize $(0$--$5)$}}}
& {\shortstack{\textbf{Reasoning}\\{\scriptsize $(0$--$5)$}}}
& {\shortstack{\textbf{Complete}\\{\scriptsize $(0$--$1)$}}}
& {\shortstack{\textbf{Queries}\\\textbf{per case}}} \\
\midrule

Oncologists
& 34 
& 0.573
& 3.000
& 2.912
& 2.676
& 1.000
& 9.412\\

GRPO Llama
& 34
&  0.537
&  2.647
&  3.206
&  2.608
&  0.931
&  4.598\\

GDPO Llama
& 34
&  0.600
&  2.931
&  3.382
&  2.804
&  0.980
&  11.275\\

GPT-5.2
& 34
&  0.544
&  2.971
&  2.951
&  3.000
&  0.873
&  37.824\\

\bottomrule
\end{tabular}
\end{table}

\section{Discussion \& Conclusion}

Clinical treatment planning requires decision-making under partial observability, yet existing clinical language model benchmarks and reinforcement learning environments often assume complete information, evaluate fixed policies, or focus primarily on diagnostic endpoints. In this work, we formulate oncology treatment planning as an interactive reinforcement learning problem and construct an environment from retrospective clinical records in which an agent must acquire patient-specific information before making a treatment recommendation, with a process-aware reward evaluating information gathering, treatment selection, and clinical reasoning. Reinforcement learning withing this environment produces a specialized Llama-3.1-8B policy that attains higher overall treatment-planning reward than both its base policy and GPT-5.2, while improving across all three clinical reward axes and learning to acquire information more selectively. Overall, these results demonstrate that retrospective clinical data can support interactive reinforcement-learning environments that provide sufficient learning signal to improve sequential clinical decision-making, even when both the patient state and the underlying clinical decision process are only partially observed and the reward is only partially verifiable. 

% Treatment planning does not provide the deterministic verification available in domains such as mathematics or programming. The documented treatment represents a clinically realized decision rather than a unique ground-truth action, while information acquisition and clinical reasoning generally lack stepwise annotations specifying the correct action at each turn. Nevertheless, our results show that partially verifiable reward can provide a useful reinforcement-learning signal and substantially improves the clinical policy over its base policy across all three reward axes. Our reward decomposes evaluation into information gathering, treatment selection, and reasoning, allowing clinically different aspects of a trajectory to be evaluated separately across clinically distinct trajectories even when no single trajectory can be labeled as uniquely correct. These results suggest that clinical RL may not require deterministic verification of the complete trajectory, provided that the reward supplies sufficient information to distinguish clinically meaningful differences between trajectories.

Formulating treatment planning as an interactive problem allows us to decompose the expert reasoning process into several distinct phases with observables. In the information acquisition phase, the policy determines which information to seek from the partially observed environment. The observable behaviors then include the number of queries and the clinical fields retrieved, and the sequence in which information is acquired. Subsequently, the policy uses the retrieved evidence to determine whether to perform further investigation or whether sufficient information is available to make a treatment recommendation. We refer to this iterative process of information acquisition, inference, and forecasting as expert reasoning. The decomposition of rewards into information gathering, treatment, and reasoning in our environment allows us to observe these distinctions between policies.

The training dynamics further suggest that reinforcement learning changes the underlying clinical policy rather than only improving the terminal treatment recommendation. The GRPO-trained Llama policy improves its clinical reward while reducing the number of queries it issues, consistent with learning a more selective information-acquisition policy rather than merely increasing the volume of retrieved information (see Table \ref{tab:query_behavior}). At the same time, the different reward axes improve at different rates, with treatment reward increasing more rapidly than information-gathering and reasoning rewards. Together, these observations suggest that interactive clinical RL environments make it possible to study how a base policy gradually adapts toward expert reasoning during training.

Beyond optimizing a single clinical policy, representing clinical decision-making as an observable trajectory creates an opportunity for studying variation among expert reasoning policies. Two policies can arrive at the same treatment recommendation while implementing meaningfully different information-acquisition and reasoning trajectories. Conversely, a policy may acquire appropriate evidence and reason coherently yet ultimately select a different but clinically defensible treatment. The absence of a unique treatment trajectory also has implications for how clinical reasoning should be modeled. 

A retrospective dataset aggregated across clinicians do not necessarily represent samples from a single expert policy. Instead, it may represent a mixture of distinct clinician practice phenotypes \cite{Krell2015-on}. We use clinician practice phenotype to denote a recurrent pattern in how a clinician or clinician subgroup acquires information, interprets clinical evidence, and selects treatment actions across comparable clinical states. For example, clinicians may differ systematically in how much evidence they acquire before committing to treatment, or in how they trade treatment intensity against toxicity for clinically similar patients. Therefore, increasing the size of a clinical dataset can increase not only the number of patient states observed but also the number of expert reasoning policies represented in the data. A clinical policy trained such trajectories may consequently encounter heterogeneous reasoning patterns from clinicians with different but clinically defensible policies. As a preliminary analysis of physician-associated variation detailed in Appendix \ref{app:behavior}, we find that the RL-trained policy does not show detectable heterogeneity across the chosen physician-associated subgroups, still, the results do not establish clinician practice phenotypes.

However, physician identity alone is insufficient to establish a clinician practice phenotype. Because each oncologist treated a different set of patients, differences between physician subgroups remain confounded by patient case mix and cannot be attributed specifically to differences in clinical decision policies. Establishing clinician practice phenotypes would require comparing information acquisition, reasoning, and treatment actions across comparable clinical states while accounting for differences in the patients encountered by each clinician. The trajectory structure captured by the environment provides the observables required for such an analysis.

\paragraph{Limitations and Future Work.} Due to resource constraint, we select the deterministic mediator based on an embedding model as opposed to generative language models. The deterministic mediator cannot fully account for the variability in realistic clinical communications, posing a limitation on how faithful the environment capture the nature of the clinical interaction. The patient representations reconstructed from the clinical records are an approximation of the actual patient encounters specifically with gastrointestinal medical oncologists. In real world, the complexity of comprehensive patient records can span across temporality, conditions, specialties, hospital systems, and so forth. Additionally, we only conduct a single RL training run per training configuration due to resource constraint. We will continue working on identifying and modeling specific phenotypes of clinical practice, extending the clinical tasks, including treatment modifications, beyond first-line treatment selection, and improving the fidelity of the patient representation setup to the realistic clinical environment. We will also keep improving the performance of the mediator by, for example, assigning the mediator an adversarial objective and co-training it and turning the protocol into a two-player stochastic game.



\bibliographystyle{unsrt}  
\bibliography{reference}

\newpage
\appendix

\section{Data \& Cohort Construction} \label{App:Data}
We conducted this study using retrospective, longitudinal, de-identified electronic health records (EHR)s from the University of California, San Francisco, integrating data from the UCSF Clinical Data Warehouse and the Observational Medical Outcomes Partnership (OMOP) Common Data Model (CDM).
Patients were first qualified using cancer-specific SNOMED CT and ICD-10-CM diagnosis codes for colon cancer in the OMOP datasets. The index date was defined as the earliest qualifying cancer diagnosis date. Adult patients with a qualifying cancer diagnosis were then linked to chemotherapy episodes recorded in the OMOP database using the relevant chemotherapy episode concept.  This is to ensure we select patients that were receiving the entirety of their oncology care within the UCSF system, as opposed to seeing UCSF for consultation but receiving care elsewhere, which would make definitive treatment planning more difficult to ascertain.

We consider an episode eligible for subsequent note collection if it was linked to a patient in the cancer cohort, had the predefined concept code for the cancer registry, and had an episode source concept name indicating chemotherapy.
Among patients with eligible chemotherapy episodes, we collected unstructured clinical notes documented from 60 days before the index date through the chemotherapy episode date and retained unstructured notes from GI Medical Oncology encounters associated with clinicians whose title was MD. 

The resulting notes were first processed into chronologically concatenated notes. These concatenated notes were then processed by GPT-5.2 to extract the treatment plan administered by the oncologist, the date on which the plan was documented, the clinical-trial enrollment, the rationale for the treatment decision, and the clinical factors informing the plan, including patient status, imaging, laboratory findings, and biomarkers. This processing extracts two classes of treatment plans, those which are eligible and those which are ineligible. The prompt for extracting the treatment plan and the reasoning behind the treatment plan is specified in Appendix \ref{app:ext_reasoning}.
We consider treatment plans eligible if first, a definitive treatment was documented. By definitive, we mean that even though there may be many clinically defensible treatment plans, the clinician selected a single treatment plan based on their own latent reasoning.  Second, the extracted treatment plan should have explicit documentation of the medication or treatment regimen. Lastly, the patient was not identified as being enrolled in a clinical trial, where decision would be based on completely unidentifiable variables, such as trial and slot availability at time of encounter.  After an eligible treatment plan was identified and filtered for,  we reconstructed the patient's clinical trajectory at the time of treatment planning. We implemented this reconstruction by first collecting GI Medical Oncology, MD-authored notes documented from 60 days before, including the treatment-plan date and then we used GPT 5.2 to process the per-patient concatenation of the notes to generate a structured record which we call the patient representation.

Clinical observations were therefore irregularly sampled according to the timing and frequency of EHR documentation, without temporal resampling or interpolation. Demographic characteristics and longitudinal diagnosis history were incorporated into the resulting  patient representation. Missing values were retained without imputation to reflect nature of the EHR. No explicit administrative or clinical censoring was applied; the observation windows were defined relative to the cancer index date, chemotherapy episode date, and treatment-plan date.

The patient representation is a structured dictionary comprising three components (see Appendix \ref{app:patient_rep}). First, a one-line summary of the history of present illness, representing the information known at the opening of the oncology consultation and serving as the initial observation. Second, an oncologic profile, including Eastern Cooperative Oncology Group (ECOG) performance status, disease staging, histology, biomarkers, and prior therapies, and, lastly, physiological data, including body measurements, vital signs, systemic examination findings, and medication history \cite{Chiu2026-gc}. The final dataset is partitioned at the eligible patient level into disjoint training, validation, and test sets, comprising 665, 143, and 142 patients, respectively, for a total of 950 patients. 

\section{Patient Representation and Environment State} \label{app:env}
We will denote the $i$-th patient by a superscript $i$. Multiple trajectories $g$ may be sampled for the same patient $i.$ Hence, trajectory-dependent objects will additionally carry a superscript $g$. We will use $k \in\{0, \ldots, K-1\}$ for the RL environment step, $t$ for token position within an agent-generated response, and $n$ for the optimizer step. In our setup, we will call the RL environment step, a turn. We will call $K$, the budget.

We distinguish the underlying clinical state of the patient $\chi^i$, the reconstructed patient representation $\rho^i$, the internal environment state $s_k^{i, g }$, and the observation $o_k^{i,g}$. The underlying clinical state $\chi^i \in \mathcal{X}_{\text {clin }}$ denotes the complete clinical condition of patient $i$, including clinically relevant properties whether or not they were measured or documented in the EHR. No clinical time elapses during a trajectory.  This is an assumption about the simulated RL environment rather than a claim that the patient is static in the real clinical time. The agent's actions do not change the patient's physiology or generate counterfactual treatment outcomes. As a result, the nontrivial information dynamics arise from the sequential information acquisition, clinical inference and decision planning that the clinical agent takes. 

The implemented environment therefore operates on a reconstructed patient representation $\rho^i=\rho\left(\ell^i, \mathcal{R}^i\right)$ where $\ell^i$ is a one-line patient summary available at the start of the episode and $\mathcal{R}^i$ is the patient-specific EHR record which is initially hidden but available through the retrieval. The construction of the patient representation that we explicitly described in the previous section refers to the mapping $\rho$. 

To facilitate rapid RL training and minimize cost, we converted the unstructured clinical not documenting patient state and clinician decision to a structured key-value dictionary. The retrievable patient record is stored as a nested key-value dictionary. Let $\mathcal{F}^i$ denote clinical fields available for patient $i$. We write $\mathcal{R}^i=\left\{\left(\xi, v^i(\xi)\right): \xi \in \mathcal{F}^i\right\}$, where $\xi$ is a clinical field and $v^i(\xi)$ is its documented value. We use a deterministic flattening operation $\operatorname{Flat}$ recursively traverses the nested dictionary and separates each item within a list into an individual clinical field while keeping the contents of each list item together. The value $v^i(\xi)$ may be numerical or categorical and denotes the actual value that is kept by the patient reconstruction pipeline, not necessarily the information in the underlying clinical state $\chi^i$ and hence the patient representation $\rho^i$ is inherently an incomplete and noisy proxy for the underlying clinical state.
A clinical field is therefore a semantic address in the patient schema, e.g., age, performance status, histology or a specific record within the prior medical history. The value $v^i(\xi)$ may be numerical, categorical or structured and denotes the actual value that is kept by the patient reconstruction pipeline, not necessarily the information in the underlying clinical state $\chi^i$ and hence the patient representation $\rho^i$ is inherently an incomplete and noisy proxy for the underlying clinical state.

The formulation of the partially observed Markov decision process will require us to know the fully observed RL trajectory $\omega^{i,g}$ which can be constructed from the fully observed state-action history $H^{i,g}_k = (s_0,a_0,s_0,a_1,...,s_k)$, the patient representation $\rho^i$, the budget $K$ and any additional information flow $x$ that passes through the environment (parser and mediator) which the agent does not observe. The fully observed RL trajectory is then parametrized by
\begin{equation}
\omega_k^{i, g }=\left( H_k^{i, g}, \rho^i, K-k,x\right) \in \mathcal{S} \times \mathcal{A},
\end{equation}
where $x$ denotes the rest of the environment information that goes to the judge model but the agent does not observe. We further remark that we will formulate the reward model as a terminal reward that does not appear within the history $H^{i,g}_k$ until the terminal step.  

The agent, however, keeps track of the partially observed RL trajectory $\iota^{i,g}$ that it receives through observations $o^{i,g}_k$. It is constructed from the partial observation-action history $h^{i,g}_k=(o_0,a_0,o_1,a_1,..., o_k)$ through an observation map $\Omega$ which filters out the patient representation and the environment information $x$; it parametrized by 
\begin{equation}
\iota_k^{i, g}=\Omega\left(\omega_k^{i, g}\right)=\left( h_k^{i, g}, K-k\right) \in \mathcal{O}\times \mathcal{A},
\end{equation}
Both RL trajectories are subject to the initial conditions
\begin{equation}
H^{i,g}_0=  s_0^{i, g}=\left(\ell^i, \rho^i\right),\quad h_0^{i, g}=\ell^i.
\end{equation}
\section{Further Details on Agent Action Space, Action Parser and Frozen Patient-Record Mediator} \label{app:med}

\subsection{Action Space}
At the turn $k$, the clinical agent with a parameterized policy  $\pi_{\theta_n}$ conditioned on the observed transcript generates a natural-language response
\begin{equation}
 \mathcal{A} \ni a_k^{i, g }  \sim \pi_{\theta_n}\left(\cdot \mid \iota_k^{i, g }\right) .
\end{equation}
Here, $a_k^{i,g}$ denotes the complete token sequence generated at one turn. Although the immediate action may contain free-form clinical reasoning at this stage, a structured action will be extracted by the parser. 

\subsection{Action Parser}
To increase the learnability of the environment, we equip the clinical agent with a simple harness, which has a deterministic parser that maps the agent-generated response to an structured action \begin{equation} \bar{a}_k^{i, g }=\operatorname{Parse}\left(a_k^{i, g }\right) \in \mathcal{A}_{\text{investigate}} \cup \mathcal{A}_{\text{recommend}}. \end{equation} The structured action space contains two action types. It may either investigate or recommend which terminates the episode. The action schema is organized by using three distinct separate tags: \verb!<reasoning></reasoning>! for reasoning steps, \verb!<query></query>! for clinical data acquisition or \verb!<treatment_recommendation></treatment_recommendation>! for terminating the episode. If a response results in an investigation type action, it may contain multiple query tags. If no valid structured action can be extracted, then no clinical action is executed on the environment and the interaction turn is consumed. Thus, any malformed output will be penalized inherently by the environment. Each extracted query is additionally screened by deterministic regular expressions in order to suppress questions that directly ask the mediator to perform clinical inference as opposed to requesting the underlying documented variable e.g. a relevant biomarker. This restriction is imposed in order to softly define the separation between information acquisition phase and the clinical inference phase. When an investigation contains both accepted and rejected queries, accepted queries are executed independently, while corrective messages are returned for rejected queries. We will reserve $\bar{a}$ for the parsed action after its been processed by the regex.

\subsection{Frozen Patient-Record Mediator}

The mediator uses a Qwen3-Embedding-0.6B model whose parameters $\phi$  are not updated during the reinforcement learning and hence frozen. This model support query and document encoding, respectively. Before constructing the retrieval index, the nested patient record is flattened into a set of retrievable clinical fields. Each clinical field is then represented through two separate channels: a key channel containing the field path and the names of fields contained within the record, and a value channel containing the documented field content. The value-channel text is truncated to 300 characters for embedding only, while the full documented value is retained for the response.

Let us denote a natural-language string via $\ell$. For each field $\xi\in\mathcal{F}^i$, let $d_{\mathrm{key}}(\xi)$ denote its key-channel representation and $d_{\mathrm{value}}^i(\xi)$ denote its value-channel representation. If we denote the query-embedding map as $\operatorname{EmbQry}_\phi$ and the document-embedding map as $\operatorname{EmbDoc}_\phi$, then for any query $\ell$, the mediator selects
\begin{equation}
\begin{aligned}
\xi^{\star}\left(\ell;\rho^i\right)
&=
\underset{\xi \in \mathcal{F}^i}{\arg\max}
\max \Biggl\{
\frac{
\left\langle
\operatorname{EmbQry}_{\phi}(\ell),
\operatorname{EmbDoc}_{\phi}\left(d_{\mathrm{key}}(\xi)\right)
\right\rangle
}{
\left\|\operatorname{EmbQry}_{\phi}(\ell)\right\|_2
\left\|\operatorname{EmbDoc}_{\phi}\left(d_{\mathrm{key}}(\xi)\right)\right\|_2
},
\\
&\hspace{4.5cm}
\frac{
\left\langle
\operatorname{EmbQry}_{\phi}(\ell),
\operatorname{EmbDoc}_{\phi}\left(d_{\mathrm{value}}^i(\xi)\right)
\right\rangle
}{
\left\|\operatorname{EmbQry}_{\phi}(\ell)\right\|_2
\left\|\operatorname{EmbDoc}_{\phi}\left(d_{\mathrm{value}}^i(\xi)\right)\right\|_2
}
\Biggr\}.
\end{aligned}
\end{equation}
where $||\cdot||$ is the Euclidean $L_2$ norm and the value-channel similarity is excluded for fields without a documented value. The mediator $\mathcal{M}_\phi$ then returns both the selected field and its full documented value
\begin{equation}
\mathcal{M}_\phi\left(\ell ; \rho^i\right)=\left(\xi^{\star}\left(\ell ; \rho^i\right), v^i\left(\xi^{\star}\left(\ell ; \rho^i\right)\right)\right).
\end{equation}
As a result, if we have an investigation action $\bar{a}$, the queries are ordered but passed as a batch through the mediator, the corresponding observation state at the next turn is then returned as
\begin{equation}
o_{k+1}^{i, g }=\operatorname{Concatenate}\left(\left[\mathcal{M}_\phi\left(\ell ; \rho^i\right): \ell\in \left(\bar{a}_k^{i, g}\right)\right]\right) ,
\end{equation}
and the partial history is updated as
\begin{equation}
\iota_{k+1}^{i, g}=\iota_k^{i, g} \oplus\left(a_k^{i, g}, o_{k+1}^{i, g}\right).
\end{equation}

\section{Policy Optimization Algorithms} \label{app:policy}
\paragraph{GRPO} For GRPO, we optimize the clinical policy such that it maximizes the expected terminal return over the training distribution of patient representations which evaluates to
\begin{equation}
J_{\text{GRPO}}(\theta)= \mathbb{E}_{i \sim \mathcal{D}_{\text {train }}} \mathbb{E}_{\omega^{i, g } \sim \mathbb{P}_\theta} \left[r_{\text {clin }}^{i, g}\right] .
\end{equation}
Here $\mathbb{P}_\theta$ is the trajectory measure induced by the environment induced jointly by the clinical policy and the environment transition rules. 

The patient-specific group mean reward with penalty is 
\begin{equation}
\bar{r}_{\mathrm{GRPO}}^i=\frac{1}{G} \sum_{g=1}^{G} r_{\mathrm{clin}}^{i, g}.
\end{equation}
The group relative advantage for each rollout $g$ is
\begin{equation}
\widehat{\mathrm{Adv}}_{\mathrm{GRPO}}^{i, g}=\frac{r_{\mathrm{GRPO}}^{i, g}-\bar{r}_{\mathrm{GRPO}}^i}{\mathrm{std}_g\left[r_{\mathrm{GRPO}}^{i, g}\right]+\epsilon_{\mathrm{num}}}.
\end{equation}

\paragraph{GDPO} For GDPO, the relative advantage is defined as follows
\begin{equation}
\begin{aligned}
\widehat{\mathrm{Adv}}_{\mathrm{info}}^{i,g}
&=
\frac{
r_{\mathrm{info}}^{i,g}
-
\operatorname{mean}_{g'=1,\ldots,G}
\left[r_{\mathrm{info}}^{i,g'}\right]
}{
\operatorname{std}_{g'=1,\ldots,G}
\left[r_{\mathrm{info}}^{i,g'}\right]
+
\epsilon_{\mathrm{num}}
},
\\[6pt]
\widehat{\mathrm{Adv}}_{\mathrm{regimen}}^{i,g}
&=
\frac{
r_{\mathrm{regimen}}^{i,g}
-
\operatorname{mean}_{g'=1,\ldots,G}
\left[r_{\mathrm{regimen}}^{i,g'}\right]
}{
\operatorname{std}_{g'=1,\ldots,G}
\left[r_{\mathrm{regimen}}^{i,g'}\right]
+
\epsilon_{\mathrm{num}}
},
\\[6pt]
\widehat{\mathrm{Adv}}_{\mathrm{reason}}^{i,g}
&=
\frac{
r_{\mathrm{reason}}^{i,g}
-
\operatorname{mean}_{g'=1,\ldots,G}
\left[r_{\mathrm{reason}}^{i,g'}\right]
}{
\operatorname{std}_{g'=1,\ldots,G}
\left[r_{\mathrm{reason}}^{i,g'}\right]
+
\epsilon_{\mathrm{num}}
},
\\[6pt]
\widehat{\mathrm{Adv}}_{\mathrm{complete}}^{i,g}
&=
\frac{
\mathbbm{1}_c^{i,g}
-
\operatorname{mean}_{g'=1,\ldots,G}
\left[\mathbbm{1}_c^{i,g'}\right]
}{
\operatorname{std}_{g'=1,\ldots,G}
\left[\mathbbm{1}_c^{i,g'}\right]
+
\epsilon_{\mathrm{num}}
},
\\[6pt]
\widehat{\mathrm{Adv}}_{\mathrm{parse}}^{i,g}
&=
\frac{
\mathbbm{1}_p^{i,g}
-
\operatorname{mean}_{g'=1,\ldots,G}
\left[\mathbbm{1}_p^{i,g'}\right]
}{
\operatorname{std}_{g'=1,\ldots,G}
\left[\mathbbm{1}_p^{i,g'}\right]
+
\epsilon_{\mathrm{num}}
}.
\end{aligned}
\end{equation}
Equal criterion weights are used in our experiments and thus we simply sum all coefficients up

$$ \widetilde{\mathrm{Adv}}_{\mathrm{GDPO}}^{i,g} = \widehat{\mathrm{Adv}}_{\mathrm{info}}^{i,g} + \widehat{\mathrm{Adv}}_{\mathrm{regimen}}^{i,g} + \widehat{\mathrm{Adv}}_{\mathrm{reason}}^{i,g} + \widehat{\mathrm{Adv}}_{\mathrm{complete}}^{i,g} + \widehat{\mathrm{Adv}}_{\mathrm{parse}}^{i,g}. $$

GDPO also applies a second normalization over all rollouts in the optimizer batch.
\begin{equation}
\widehat{\mathrm{Adv}}_{\mathrm{GDPO}}^{i, g}=\frac{\widetilde{\mathrm{Adv}}_{\mathrm{GDPO}}^{i, g}-\operatorname{mean}_{(i, g) \in B \times\left\{1, \ldots, N_g\right\}}\left[\widetilde{\mathrm{Adv}}_{\mathrm{GDPO}}{ }^{i, g}\right]}{\operatorname{std}_{(i, g) \in B \times\left\{1, \ldots, N_g\right\}}\left[\widetilde{\mathrm{Adv}}_{\mathrm{GDPO}}\right]+\epsilon_{\mathrm{num}}} .
\end{equation}

\paragraph{Objective}
Let $x^{i,g}_{k,t}$ be the agent-generated token at environment step $k$ and token position $t$ and let $\iota_{t,k}$ be the policy context immediately before that token. Then the likelihood ratio between the current policy and the behavior-policy is

% \textbf{Remove A,E}
\begin{equation}
w_{k, t}^{i, g}(\theta)=\frac{\pi_{\theta(n) \mid A}\left(x_{k, t}^{i, g} \mid \iota_{k, t}^{i, g}\right)}{\pi_{\bar{\theta}(n) \mid A}\left(x_{k, t}^{i, g} \mid \iota_{k, t}^{i, g}\right)} .
\end{equation}
Let $\widehat{\mathrm{A d v}}^{i, g}$ denote either the GRPO coefficient or the GDPO coefficient. Because the reward is only available after termination, we broadcast this to every agent-generated token in the rollout, the corresponding clipped token-level contribution

\begin{equation}
\mathcal{L}_{k, t}^{i, g}(\theta)=\min \left[w_{k, t}^{i, g}(\theta) \widehat{\operatorname{Adv}}^{i, g}, \operatorname{clip}\left(w_{k, t}^{i, g}(\theta), 1-\varepsilon, 1+\varepsilon\right) \widehat{\mathrm{Adv}}^{i, g}\right] .
\end{equation}
Only tokens generated by the clinical agent contribute to the policy-gradient term. Tokens belonging to the system prompt, initial patient observation, parser messages, or mapper responses are masked. The clipped objective is then

\begin{equation}
J_{\text {clip }}(\theta) = \mathbb{E}\left[\frac{1}{N_g} \sum_{g=1}^{N_g} \frac{1}{N_{\text {tok }}^{i, g}} \sum_{k, t} \mathbbm{1}_{\textbf{agent}} (k, t) \mathcal{L}_{k, t}^{i, g}(\theta)-\beta_{\mathrm{KL}} D_{\mathrm{KL}}\left(\pi_{\theta(n) \mid A} \| \pi_{\mathrm{ref}}^{\mid A}\right)\right], 
\end{equation}
where the reference policy $\pi_{\text{ref}}$ is the frozen initial clinical-agent policy.

\section{Experimental Setup} 
\label{sec:experiments}

We train Llama-3.1-8B-Instruct using parameter-efficient policy optimization under a five-turn interaction budget. At each turn, the agent can generate at most 512 new tokens. We generate a group of eight trajectories for each patient. We use AdamW optimizer with a constant learning rate of $1\times 10^{-6}$. We enforce Kullback–Leibler (KL) divergence to regularize the update to the new policy against the reference policy with coefficient $0.01$. We apply low rank adaptation (LoRA) to train the agent policy. We attach LoRA adapters to all attention and feed-forward projection matrices, i.e., query, key, value, output, gate, up, and down projections, of the agent policy, while the rest of the network is frozen and loaded with 4-bit quantization. We set the LoRA rank to \(32\), scaling factor \(\alpha\) to \(32\), and dropout to \(0\).

The environment uses a frozen Qwen3-Embedding-0.6B mediator and GPT-5.2 judge throughout training and evaluation. We select GPT-5.2 because it is the most performant model available through our institutional secure endpoint, and set temperature to 0 with up to three retries. The context window is by default 400,000 tokens with maximum output 128,000 tokens per response. The judge returns a dictionary containing the numeric score for each reward axis together with natural-language grading rationale.

During the training, we use the complete trajectory to update the agent policy. The terminal reward is broadcast to all tokens generated by the agent policy for gradient update, while the tokens that are not generated by the agent, including system instructions and mediator responses, are masked from the gradient update. We assign equal weight to each reward axis during reward scalarization. To encourage diversity in the rollout, we use sampling-based decoding with temperature $0.6$ and top-$p$ $0.95$. We process each of the 665 training patients once, so training consists of 665 patient-level steps. Within each patient-level step, corresponding trajectory group may undergo multiple inner optimization steps. We checkpoint the policy every 50 patient-level steps. For each training configuration, we independently select the checkpoint with the highest Mean@1 total scalar reward on the validation set with temperature set 0 and report test-set performance using the selected checkpoint. The test set is not used for checkpoint selection.

For both GRPO and GDPO, each patient contributes one group of eight sampled trajectories, which we reuse for three inner optimization steps before proceeding to the next patient. The inner optimization steps allows reusing the already sampled trajectories for multiple gradient updates. We optimize the PPO clipped surrogate with $\epsilon = 0.2$ to bound how far each update moves the policy.

\section{Additional Figures}

\begin{figure}[h]
    \centering
    \includegraphics[width=1\linewidth]{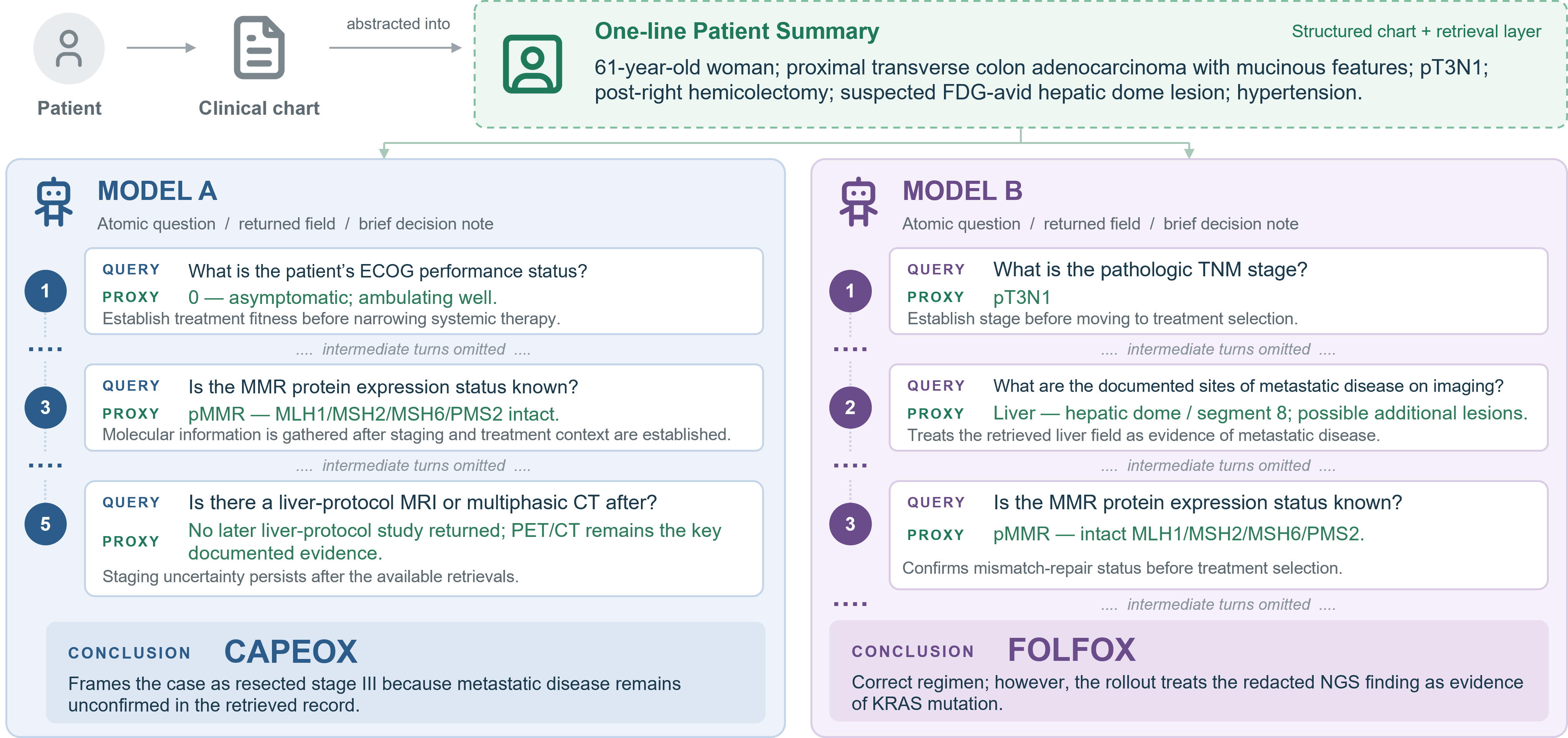}
    \caption{Illustrative example of trajectories from the clinical agent given a patient case. The clinical agent interacts the patient environment in order to retrieve relevant information necessary for selecting the treatment. In the process, they will describe the reasoning behind the retrieval certain information, and the inference on the retrieved information. }
    \label{fig:workflow}
\end{figure}
\begin{figure}[h]
    \centering
    \begin{subfigure}[h]{0.32\linewidth}
        \centering
        \includegraphics[width=\linewidth]
        {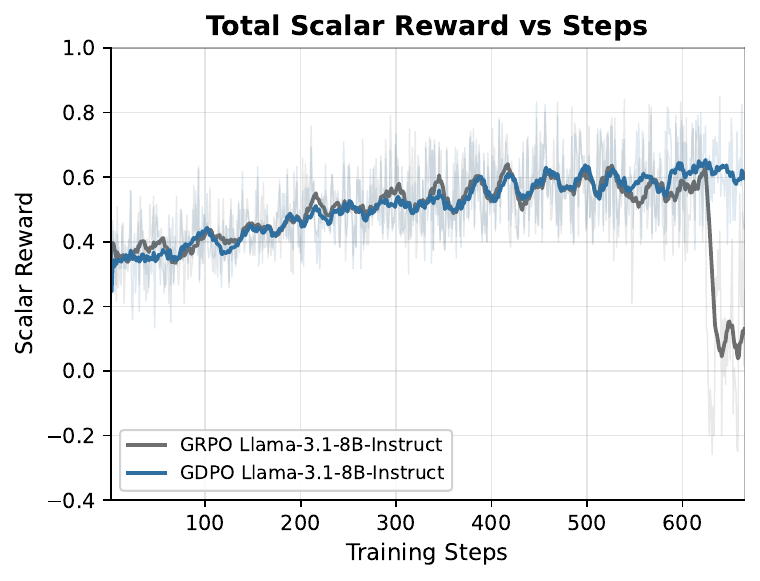}
        \caption{Total scalar reward.}
        \label{fig:learning_scalar}
    \end{subfigure}
    \hspace{0.06\linewidth}
    \begin{subfigure}[h]{0.32\linewidth}
        \centering
        \includegraphics[width=\linewidth]
        {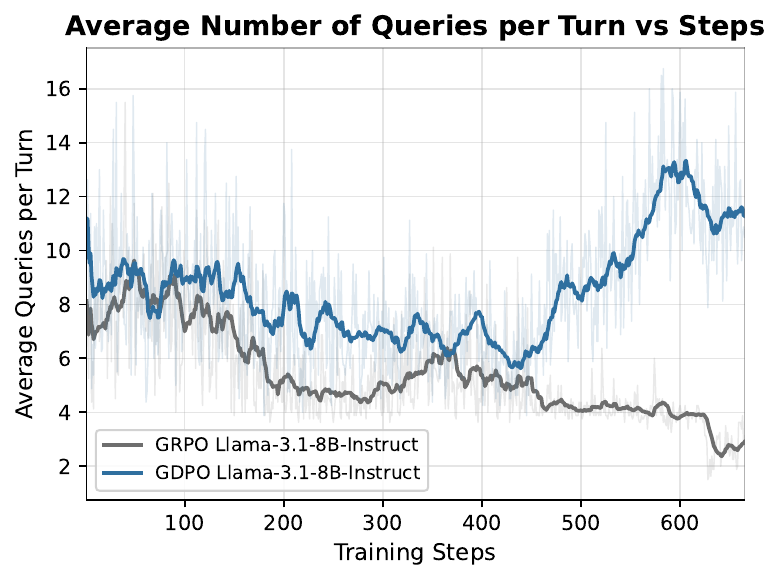}
        \caption{Average number of queries.}
        \label{fig:learning_queries}
    \end{subfigure}

    \caption{Learning dynamics of GRPO and GDPO with Llama-3.1-8B-Instruct. We report  the total scalar reward and the average number of queries issued by the policy. Dark curves show a 15-step moving average and faint curves show the corresponding unsmoothed training trajectory. Each curve corresponds to a single training run. }
    \label{fig:learning_dynamics}
\end{figure}
\section{Reinforcement Learning using Different Policy Backbones}
\begin{figure}[h]
    \centering
    % Row 1: reward axes
    \begin{subfigure}[h]{0.32\linewidth}
        \centering
        \includegraphics[width=\linewidth]
        {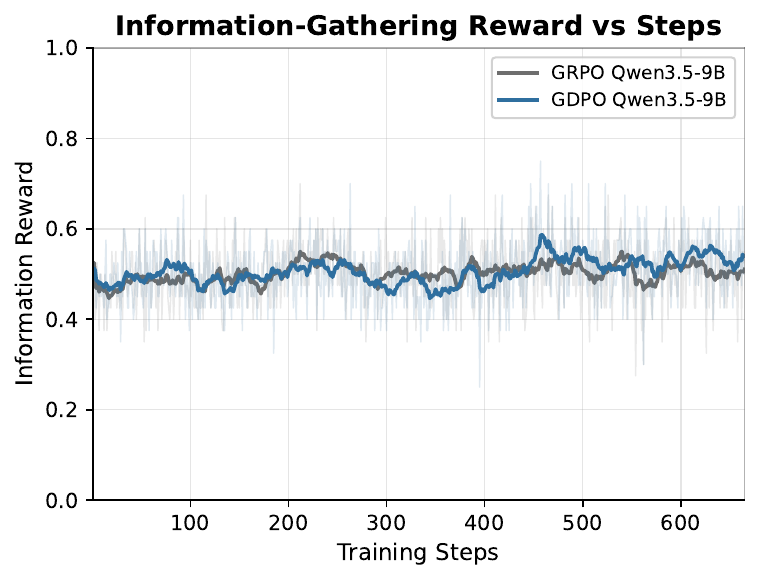}
        \caption{Information-gathering reward.}
        \label{fig:backbone_information}
    \end{subfigure}
    \hfill
    \begin{subfigure}[h]{0.32\linewidth}
        \centering
        \includegraphics[width=\linewidth]
        {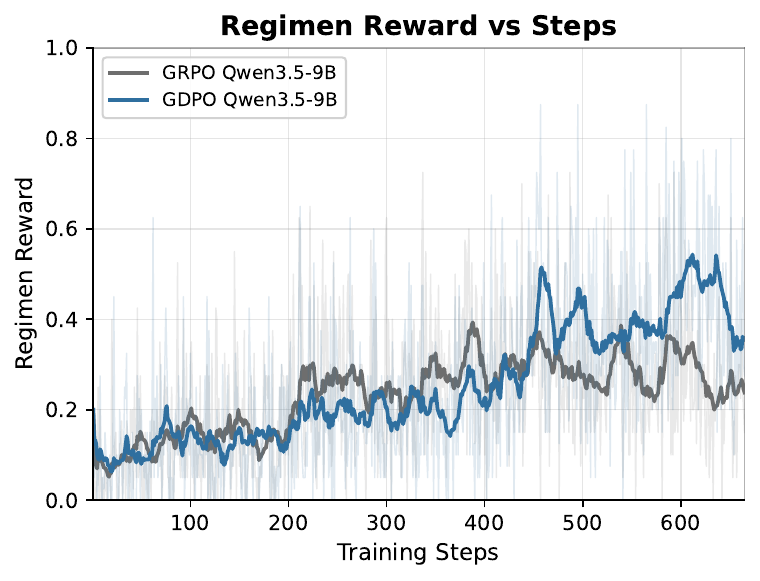}
        \caption{Treatment reward.}
        \label{fig:backbone_regimen}
    \end{subfigure}
    \hfill
    \begin{subfigure}[h]{0.32\linewidth}
        \centering
        \includegraphics[width=\linewidth]
        {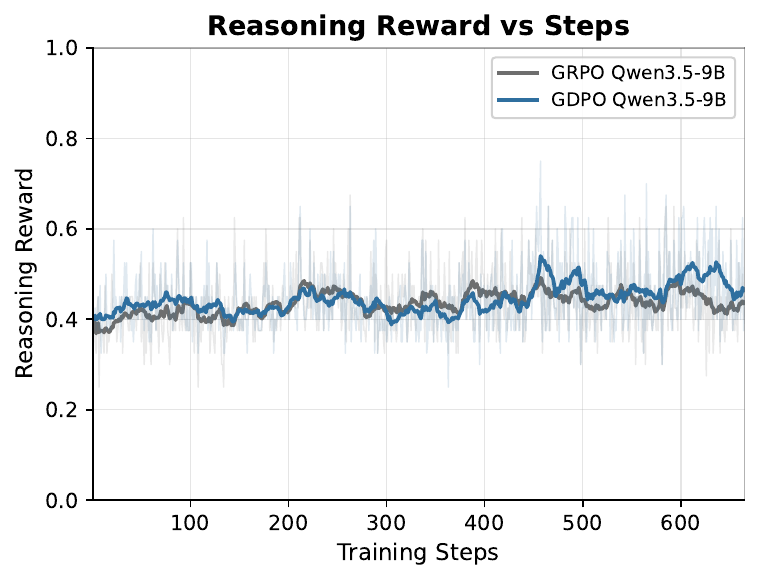}
        \caption{Reasoning reward.}
        \label{fig:backbone_reasoning}
    \end{subfigure}

    \vspace{0.6em}

    % Row 2: scalar reward + query behavior
    \begin{subfigure}[h]{0.32\linewidth}
        \centering
        \includegraphics[width=\linewidth]
        {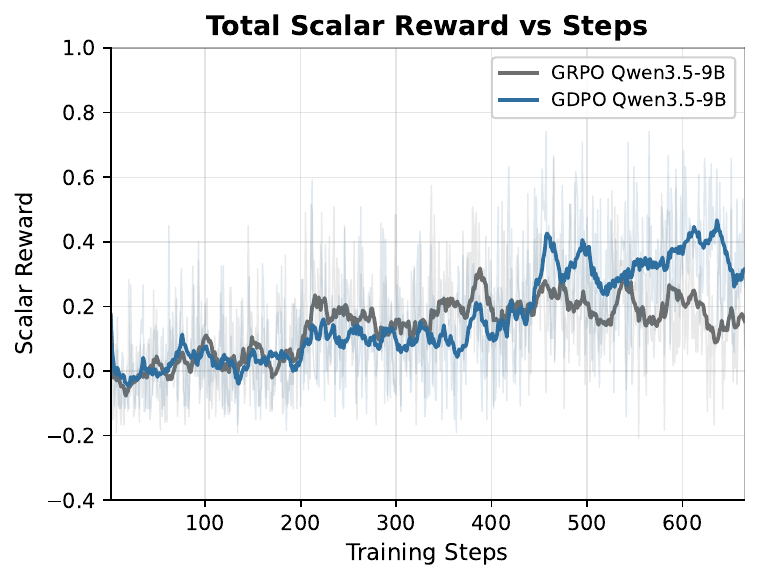}
        \caption{Total scalar reward.}
        \label{fig:backbone_scalar}
    \end{subfigure}
    \hspace{0.06\linewidth}
    \begin{subfigure}[h]{0.32\linewidth}
        \centering
        \includegraphics[width=\linewidth]
        {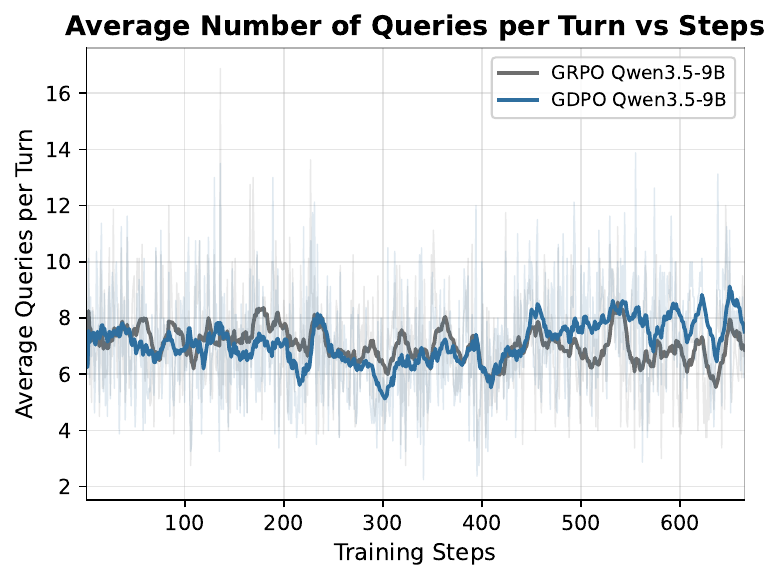}
        \caption{Average number of queries.}
        \label{fig:backbone_queries}
    \end{subfigure}

    \caption{GRPO and GDPO training dynamics on Qwen3.5-9B policy backbone. Dark curves show a 15-step moving average and faint curves show the corresponding unsmoothed training trajectory. Each curve corresponds to a single training run.}
    \label{fig:qwen training}
\end{figure}

\begin{figure}[h]
    \centering
    % Row 1: reward axes
    \begin{subfigure}[h]{0.32\linewidth}
        \centering
        \includegraphics[width=\linewidth]
        {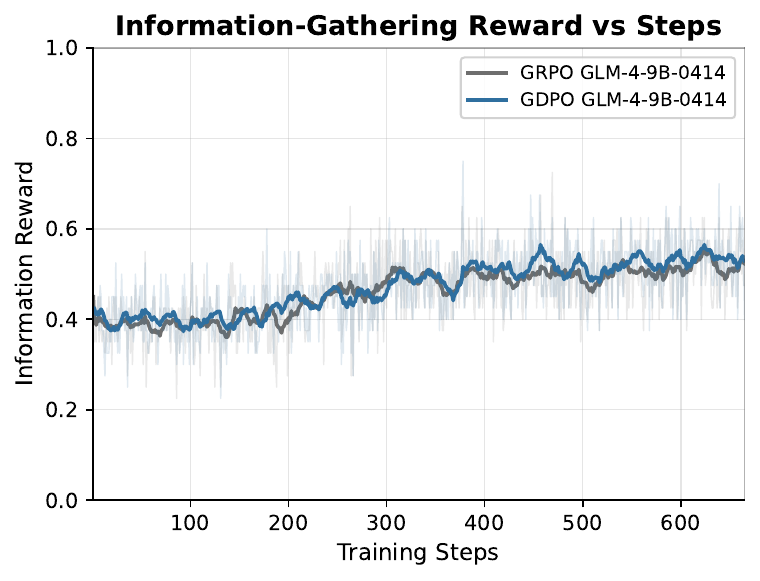}
        \caption{Information-gathering reward.}
        \label{fig:backbone_information}
    \end{subfigure}
    \hfill
    \begin{subfigure}[h]{0.32\linewidth}
        \centering
        \includegraphics[width=\linewidth]
        {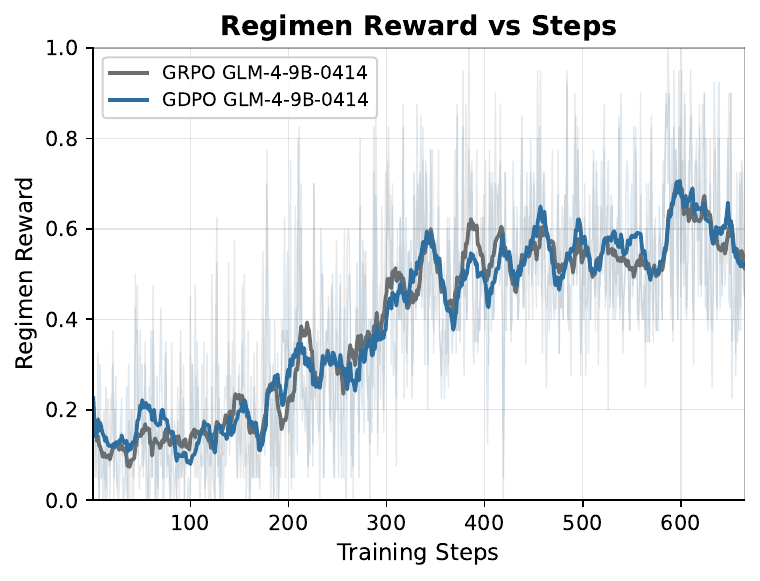}
        \caption{Treatment reward.}
        \label{fig:backbone_regimen}
    \end{subfigure}
    \hfill
    \begin{subfigure}[h]{0.32\linewidth}
        \centering
        \includegraphics[width=\linewidth]
        {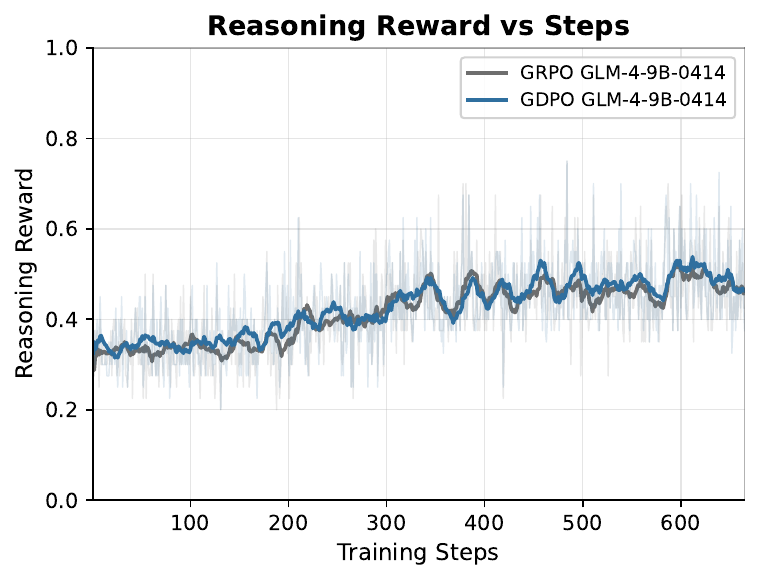}
        \caption{Reasoning reward.}
        \label{fig:backbone_reasoning}
    \end{subfigure}

    \vspace{0.6em}

    % Row 2: scalar reward + query behavior
    \begin{subfigure}[h]{0.32\linewidth}
        \centering
        \includegraphics[width=\linewidth]
        {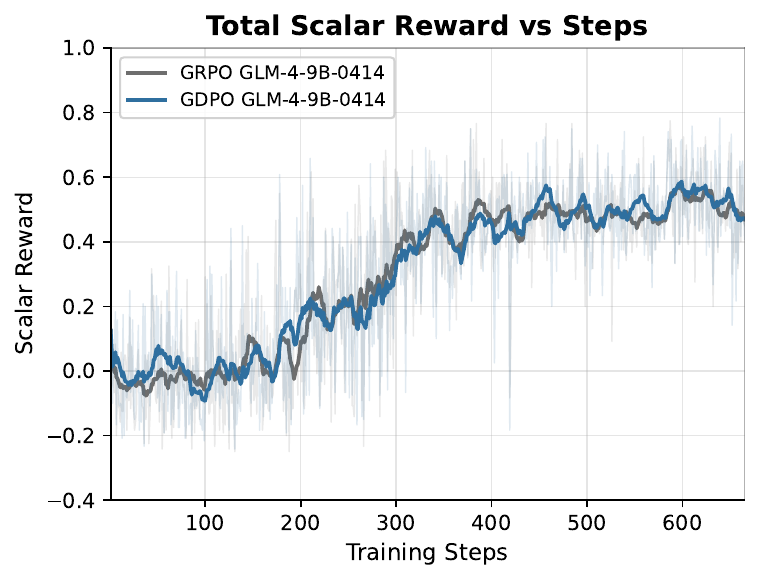}
        \caption{Total scalar reward.}
        \label{fig:backbone_scalar}
    \end{subfigure}
    \hspace{0.06\linewidth}
    \begin{subfigure}[h]{0.32\linewidth}
        \centering
        \includegraphics[width=\linewidth]
        {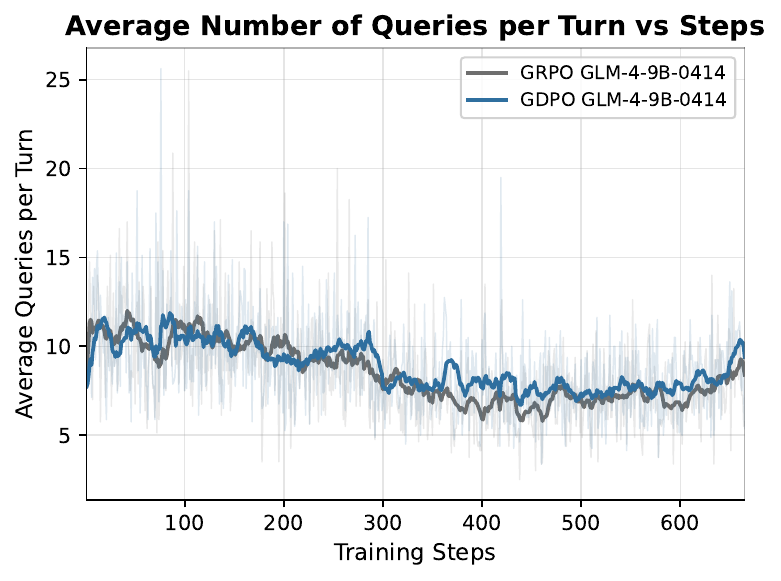}
        \caption{Average number of queries.}
        \label{fig:backbone_queries}
    \end{subfigure}

    \caption{GRPO and GDPO training dynamics on GLM-4-9B-0414 policy backbone. Dark curves show a 15-step moving average and faint curves show the corresponding unsmoothed training trajectory. Each curve corresponds to a single training run.}
    \label{fig:glm training}
\end{figure}

We investigate the efficacy of the reinforcement learning (RL) environment setup across different policy backbones, including Llama-3.1-8B-Instruct, Qwen-3.5 9B and GLM-4-9B-0414. RL improves performance on different policy backbones, although the magnitude of improvement depends strongly on the backbone. For Qwen3.5-9B, GRPO and GDPO increase Mean@3 from \(-0.0142\) for the base policy to \(0.1994\) and \(0.3429\), respectively. For GLM-4-9B-0414, the Mean@3 improves from $-0.0163$ to $0.5313$ and $0.5105$ after GRPO and GDPO training, respectively. Both policy backbones demonstrate improvement in each reward axes during training, as shown in Figures \ref{fig:qwen training} and \ref{fig:glm training}. 
\begin{table}[h]
\centering
\caption{
Performance on each reward axis.
Information, treatment, and reasoning are scored from 0--5;
completeness and format are scored from 0--1. Best results are shown in \textbf{bold} and second-best results are \underline{underlined}.
}
\label{tab:reward_axes}
\small
\setlength{\tabcolsep}{4.5pt}
\renewcommand{\arraystretch}{1.08}

\begin{tabular}{
    ll
    S[table-format=1.3]
    S[table-format=1.3]
    S[table-format=1.3]
    S[table-format=1.3]
    S[table-format=1.3]
}
\toprule
 
&
& {\shortstack{Information\\{\scriptsize $(0$--$5)$}}}
& {\shortstack{Regimen\\{\scriptsize $(0$--$5)$}}}
& {\shortstack{Reasoning\\{\scriptsize $(0$--$5)$}}}
& {\shortstack{Complete\\{\scriptsize $(0$--$1)$}}}
& {\shortstack{Format\\{\scriptsize $(0$--$1)$}}} \\
\midrule

GRPO Llama-3.1-8B-Instruct & Mean@3 & 2.754 & \underline{3.366} & 2.688 & 0.960 & 1.000 \\
 & Best@3 & 3.000 & \underline{3.789} & 2.944 & 1.000 & 1.000 \\
GDPO Llama-3.1-8B-Instruct & Mean@3 & \underline{2.953} & \textbf{3.453} & \underline{2.791} & \textbf{0.988} & 1.000 \\
 & Best@3 & \underline{3.176} & \textbf{3.845} & \underline{3.077} & 1.000 & 1.000 \\
GRPO Qwen3.5-9B & Mean@3 & 2.606 & 1.453 & 2.298 & 0.439 & 1.000 \\
 & Best@3 & 3.007 & 2.796 & 2.768 & 0.824 & 1.000 \\
GDPO Qwen3.5-9B & Mean@3 & 2.725 & 2.033 & 2.399 & 0.664 & 1.000 \\
 & Best@3 & 3.063 & 3.183 & 2.852 & 0.951 & 1.000 \\
GRPO GLM-4-9B-0414 & Mean@3 & 2.728 & 2.967 & 2.444 & \underline{0.972} & 1.000 \\
 & Best@3 & 3.056 & 3.556 & 2.859 & 0.993 & 1.000 \\
GDPO GLM-4-9B-0414 & Mean@3 & 2.629 & 2.854 & 2.455 & 0.953 & 1.000 \\
 & Best@3 & 2.958 & 3.521 & 2.901 & 1.000 & 1.000 \\
GPT-5.2 & Mean@3 & \textbf{2.977} & 2.702 & \textbf{3.016} & 0.831 & 1.000 \\
 & Best@3 & \textbf{3.232} & 3.380 & \textbf{3.352} & 0.972 & 1.000 \\
Llama-3.1-8B-Instruct & Mean@3 & 2.124 & 1.876 & 1.927 & 0.859 & 1.000 \\
 & Best@3 & 2.458 & 2.746 & 2.310 & 1.000 & 1.000 \\
Qwen3.5-9B & Mean@3 & 2.406 & 0.437 & 2.085 & 0.143 & 1.000 \\
 & Best@3 & 2.852 & 1.063 & 2.338 & 0.338 & 1.000 \\
GLM-4-9B-0414 & Mean@3 & 2.002 & 0.678 & 1.723 & 0.225 & 1.000 \\
 & Best@3 & 2.394 & 1.542 & 2.070 & 0.500 & 1.000 \\
\addlinespace[2pt]

\bottomrule
\end{tabular}
\end{table}


\section{Quantitative Analysis of Information-Acquisition Behavior} \label{app:behavior}
We quantify the behavior change in information acquisition before and after reinforcement learning and the difference between different policies. Using the trajectories collected from running the held-out test set ($n=142$), we extract the number of queries the policy issues, the number of unique clinical field retrieved by the mediator, the number of terminal turn, the numeric score for information-gathering on a scale from 0 to 5, and calculate the rate of redundant queries. We first average over 3 sampled trajectories within each patient, and then conduct the statistical comparison between policies in pair per patient. We report the mean paired different with a $95\%$ bootstrap confidence interval, the median, and a paired Wilcoxon signed-rank $p$.

Compared to base Llama, GRPO reduces the number of queries by $4.65$ per case and the number of unique fields retrieved by $2.09$, while increasing information-gathering reward by $0.63$. Conversely, GDPO increases the number of queries by $2.22$ per case and the number of unique fields retrieved by $2.65$, while increasing information-gathering reward by $0.83$. Compared to GPT-5.2, both RL-trained Llama polices issue approximately $27$ to $34$ fewer queries per case and retrieve approximately $10$ to $14$ fewer unique fields than GPT-5.2.

Together, these results provide quantitative evidence that the learned policy adapts its interaction behavior to the structure of the reinforcement learning environment. The information-gathering reward rubric favors acquisition of variables relevant to the decision while penalizing unnecessary or redundant investigation, including continued information gathering after sufficient evidence is available. 
\begin{table}[h]
\centering
\caption{
Patient-level changes in information-gathering behavior on the held-out test set. Values report the mean paired difference between each RL-trained policy based on Llama-3.1-8B-Instruct and the reference policy, with 95\% bootstrap confidence intervals.
}
\label{tab:query_behavior}
\resizebox{\textwidth}{!}{
\begin{tabular}{llrrr}
\toprule
Policy & Queries/case & Unique fields & Information Reward \\
\midrule
Llama-3.1-8B (Base) & $8.98$ $[8.42,9.57]$ & $6.21$ & $2.12$ \\
GRPO-Llama & $4.33$ $[4.17,4.51]$ & $4.12$ & $2.75$ \\
GDPO-Llama & $11.20$ $[10.84,11.57]$ & $8.86$ & $2.95$ \\
Qwen (Base) & $7.26$ $[6.87,7.65]$ & $5.20$ & $2.41$ \\
GRPO-Qwen & $6.96$ $[6.66,7.25]$ & $5.18$ & $2.61$ \\
GDPO-Qwen & $7.96$ $[7.57,8.35]$ & $5.73$ & $2.73$ \\
GLM-4-9B (Base) & $10.40$ $[9.81,11.03]$ & $5.23$ & $2.00$ \\
GRPO-GLM & $8.07$ $[7.69,8.47]$ & $6.12$ & $2.73$ \\
GDPO-GLM & $8.19$ $[7.76,8.64]$ & $6.04$ & $2.63$ \\
GPT-5.2 & $38.41$ $[36.57,40.27]$ & $18.98$ & $2.98$ \\

\bottomrule
\end{tabular}
}
\end{table}
\section{Exploratory Analysis Across Primary Oncologist Subgroups} 
To examine whether policy performance is consistent across physician-associated patient subgroups, we stratified the held-out test set according to the patient's primary oncologist. The primary oncologist was identified from the author of the clinical note from which the documented treatment plan was extracted. We selected the three oncologists with the largest number of patients in the test set, corresponding to 47, 33, and 23 patients, respectively. We then compared the per-patient reward distributions of the evaluated policies within each subgroup. 

\begin{figure}[t]
    \centering

    \begin{subfigure}[t]{0.63\textwidth}
        \vspace{0pt}
        \centering
        \includegraphics[width=\linewidth]{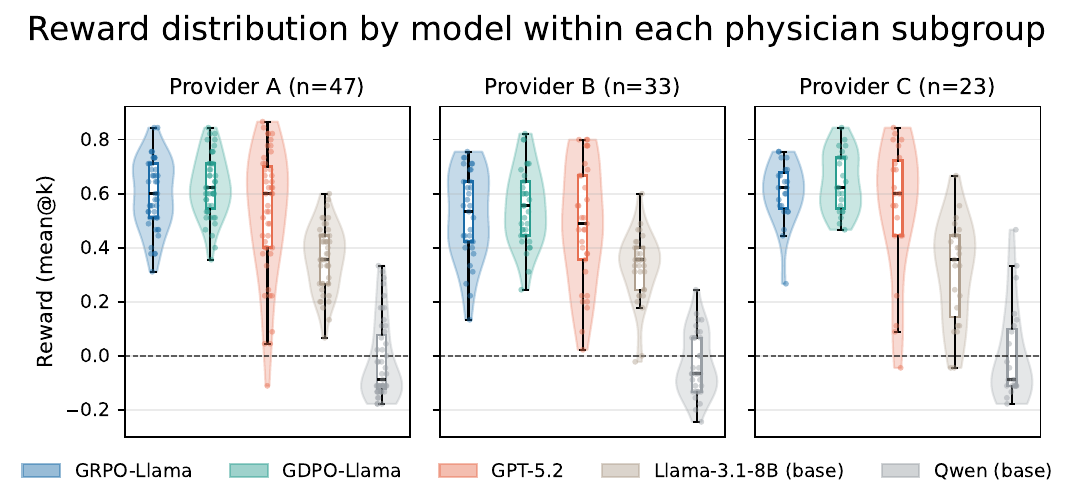}
        \caption{Per-patient reward distributions across policies within the three largest primary-oncologist subgroups.}
        \label{fig:provider_reward_distribution}
    \end{subfigure}
    \hfill
    \begin{subfigure}[t]{0.33\textwidth}
        \vspace{0pt}
        \centering
        \includegraphics[width=\linewidth]{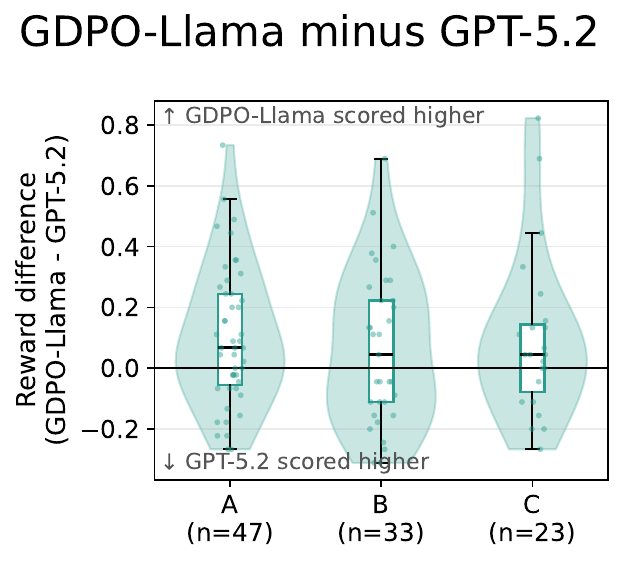}
        \caption{Paired per-patient reward difference between the specialized GDPO policy and GPT-5.2.}
        \label{fig:provider_gdpo_advantage}
    \end{subfigure}

    \caption{
    Policy performance across primary-oncologist subgroups.
    \textbf{(a)} Per-patient reward distributions for each policy.
    \textbf{(b)} Per-patient reward difference between the specialized GDPO policy and GPT-5.2. Positive values indicate higher reward for GDPO.
    }
    \label{fig:provider_subgroup_analysis}
\end{figure}

The GDPO-trained Llama policy has a positive median reward difference relative to GPT-5.2 in all three physician subgroups, with median paired differences of $0.067$, $0.044$, and $0.044$ for Providers A, B, and C, respectively (see Figure~\ref{fig:provider_subgroup_analysis} and Table~\ref{tab:provider_paired_comparison}). The difference reached statistical significance in Provider A (mean paired difference $0.100$, $95\%$ CI $[0.036, 0.169]$, Holm-corrected $p=0.018$) but not in Providers B or C (Holm-corrected $p=0.396$ and $p=0.422$, respectively). We found no evidence that the GDPO--GPT-5.2 reward difference varied across physician subgroups (Kruskal--Wallis $H=0.464$, $p=0.793$). Absolute reward distributions varied across physician subgroups for the GDPO policy (Kruskal--Wallis $H=8.585$, $p=0.014$; see Table~\ref{tab:provider_raw_reward}), although these comparisons are confounded by differences in patient case mix. 

\begin{table*}[t]
\centering
\caption{
Paired comparisons of policy reward against GPT-5.2 within physician-associated patient
subgroups. Differences are computed per patient using Mean@3 scalar reward. Confidence intervals
are 95\% bootstrap confidence intervals for the mean paired difference. \(p\)-values are from paired
Wilcoxon signed-rank tests with Holm correction within each provider subgroup.
}
\label{tab:provider_paired_comparison}
\small

\begin{tabular}{llrrcrr}
\toprule
Provider & Policy & \(n\) & Mean \(\Delta\) & 95\% CI &
Median \(\Delta\) & Holm \(p\) \\
\midrule

\multirow{4}{*}{A}
& GRPO Llama  & 47 & 0.071 & [0.009, 0.137] & 0.044 & 0.109 \\
& GDPO Llama  & 47 & 0.100 & [0.036, 0.169] & 0.067 & \textbf{0.018} \\
& Llama base  & 47 & -0.165 & [-0.239, -0.087] & -0.200 & \textbf{\(7.6\times10^{-4}\)} \\
& Qwen base   & 47 & -0.544 & [-0.619, -0.463] & -0.556 & \textbf{\(1.6\times10^{-8}\)} \\

\midrule

\multirow{4}{*}{B}
& GRPO Llama  & 33 & 0.032 & [-0.055, 0.124] & 0.000 & 0.674 \\
& GDPO Llama  & 33 & 0.063 & [-0.015, 0.149] & 0.044 & 0.396 \\
& Llama base  & 33 & -0.158 & [-0.231, -0.082] & -0.133 & \textbf{0.002} \\
& Qwen base   & 33 & -0.519 & [-0.593, -0.444] & -0.556 & \textbf{\(2.2\times10^{-6}\)} \\

\midrule

\multirow{4}{*}{C}
& GRPO Llama  & 23 & 0.055 & [-0.045, 0.165] & 0.044 & 0.770 \\
& GDPO Llama  & 23 & 0.096 & [-0.002, 0.210] & 0.044 & 0.422 \\
& Llama base  & 23 & -0.238 & [-0.375, -0.092] & -0.178 & \textbf{0.014} \\
& Qwen base   & 23 & -0.547 & [-0.664, -0.421] & -0.622 & \textbf{\(1.6\times10^{-4}\)} \\

\bottomrule
\end{tabular}

\vspace{0.5em}
\raggedright
\footnotesize
\(\Delta\) denotes policy reward minus GPT-5.2 reward.
Bold \(p\)-values indicate \(p<0.05\) after Holm correction.
\end{table*}
\begin{table}[t]
\centering
\caption{
Per-patient Mean@3 scalar reward across physician-associated patient subgroups.
Kruskal--Wallis comparisons are descriptive because provider subgroups may differ in patient case mix.
}
\label{tab:provider_raw_reward}
\small
\begin{tabular}{lrrrrr}
\toprule
Policy &
Provider A &
Provider B &
Provider C &
\(H\) &
\(p\) \\
\midrule

GRPO Llama
& 0.599 & 0.519 & 0.604 & 5.400 & 0.067 \\

GDPO Llama
& 0.627 & 0.550 & 0.644 & 8.585 & \textbf{0.014} \\

GPT-5.2
& 0.528 & 0.487 & 0.549 & 1.784 & 0.410 \\

Llama base
& 0.363 & 0.329 & 0.311 & 1.426 & 0.490 \\

Qwen base
& -0.017 & -0.032 & 0.002 & 0.248 & 0.883 \\

\bottomrule
\end{tabular}
\end{table}
\FloatBarrier
\section{Prompt Details} \label{app:prompt}
\subsection{Instruction for extracting clinical decision making process}  \label{app:ext_reasoning}
\promptfile{Clinician Decision Extraction Prompt}{treatment_plan_instruction.tex}

\subsection{Instruction for building the patient representation}\label{app:patient_rep}
\promptfile{Patient Representation Prompt}{patient_representation.tex}
\subsection{Instruction for the judge model}

\promptfile{Judge Model Prompt}{rubric_instruction.tex}

\subsection{Instruction for the clinical agent}

\promptfile{Clinical Agent Prompt}{agent_instruction.tex}
\FloatBarrier

\end{document}